%% file: main_ai4good_workshop.tex
\documentclass{article}

\usepackage{microtype}
\usepackage{graphicx}
\usepackage{subcaption}
\usepackage{booktabs} 

\usepackage{hyperref}

\usepackage[accepted]{icml2026}

\usepackage{amsmath}
\usepackage{amssymb}
\usepackage{mathtools}
\usepackage{amsthm}

\theoremstyle{plain}
\newtheorem{theorem}{Theorem}[section]

\theoremstyle{definition}
\newtheorem{definition}[theorem]{Definition}

\theoremstyle{remark}

\usepackage[textsize=tiny]{todonotes}

\usepackage[dvipsnames]{xcolor}

\usepackage{lipsum}
\usepackage{enumitem}
\usepackage{array}

\usepackage[capitalize,noabbrev,nameinlink]{cleveref}

\definecolor{cornflowerblue}{rgb}{0.39, 0.58, 0.93}
\definecolor{cornflowerscomplement}{rgb}{0.93, 0.74, 0.39}
\definecolor{cornflowersanalogouspurple}{rgb}{0.47, 0.39, 0.93}

\hypersetup{
    colorlinks=true,
    linkcolor=cornflowerblue,
    filecolor=magenta,      
    urlcolor=cornflowersanalogouspurple,
    citecolor=cornflowerblue,
    pdftitle={Watermarking for Proprietary Dataset Protection},
    pdfpagemode=FullScreen,
}

\newcommand\checkemoji[1]{\includegraphics[width=0.3cm]{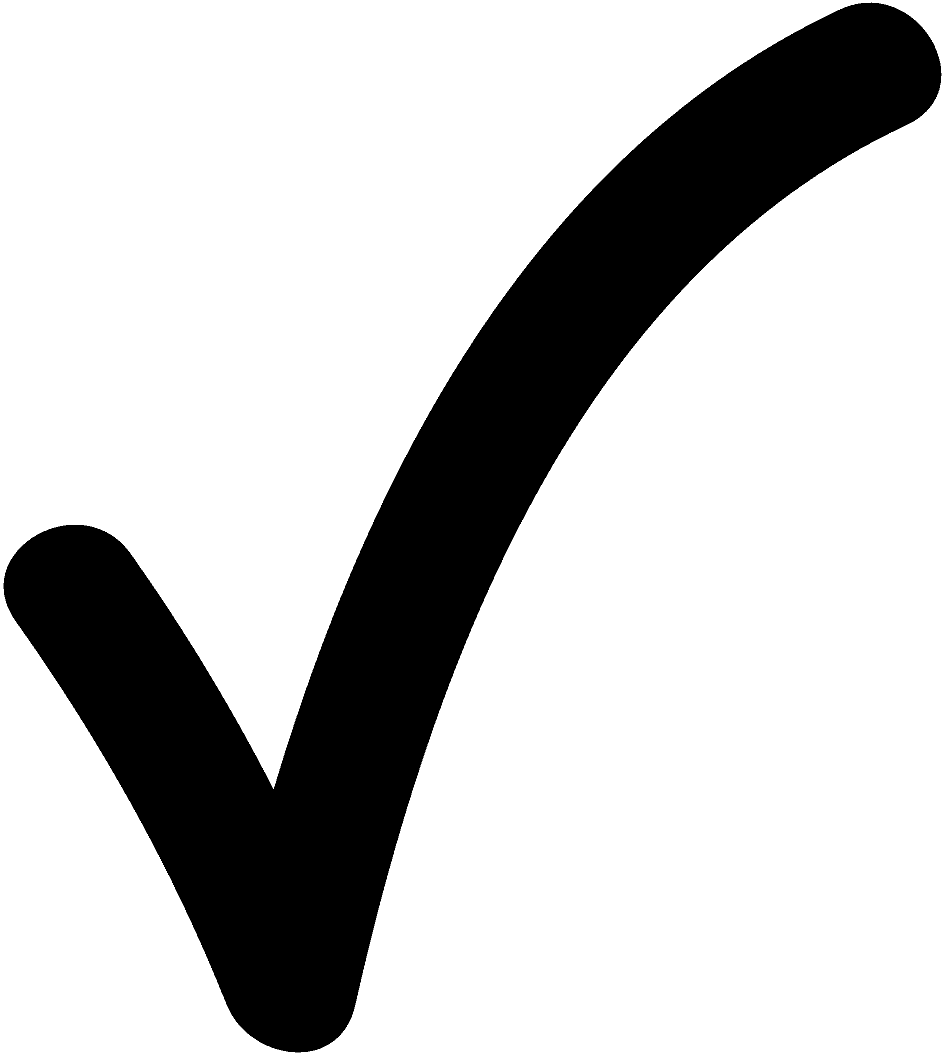}}

\icmltitlerunning{Watermarking for Proprietary Dataset Protection}

\begin{document}

\twocolumn[
  \icmltitle{Watermarking for Proprietary Dataset Protection}



  \icmlsetsymbol{equal}{*}

  \begin{icmlauthorlist}
    \icmlauthor{John Kirchenbauer}{umd}
    \icmlauthor{Brian R. Bartoldson}{llnl}
    \icmlauthor{Bhavya Kailkhura}{llnl}
    \icmlauthor{Tom Goldstein}{umd}
  \end{icmlauthorlist}

  \icmlaffiliation{umd}{University of Maryland}
  \icmlaffiliation{llnl}{Lawrence Livermore National Labs}

  \icmlcorrespondingauthor{John Kirchenbauer}{jkirchen@umd.edu}


  \vskip 0.3in
]



\printAffiliationsAndNotice{}  

\begin{abstract}
A growing body of literature suggests that training data membership inference problems are fundamentally hard tasks in modern language modeling settings. We argue that output watermarking techniques are the right gadget to make training membership tests for generative models more tractable, based on prior results showing that language models exhibit residual watermark ``radioactivity'' under partially watermarked training datasets. We pit a watermark-based dataset inference approach head-to-head against traditional loss-based membership inference methods and show that watermarking can achieve comparable membership detection performance when subset exposure is high enough, under an alternate set of assumptions.
\end{abstract}

\section{Introduction}

Modern language models perform complex knowledge work of growing economic value, but the regulatory frameworks governing fair use of the web-scraped data they train on remain underdeveloped. Recent litigation suggests content owners like news websites and independent authors may be entitled to compensation for inclusion of their datasets in large-scale AI model training. Answering such data-use questions in high-stakes settings requires a concrete definition of what it means to test whether some data was included in a model's training dataset.

While the fully general training data attribution problem asks how a model's test-time behaviors are caused by specific training instances, the question at hand in contemporary fair-use deliberations is actually just \textit{membership}. Membership inference attacks (MIAs) ask whether a specific sample was in a model's training dataset; dataset inference attacks (DIAs) generalize this to whole collections. As the more relevant setting for IP and generative-model training disputes, in this work we study the DIA setup but refer to both problem types collectively as membership problems.

\begin{figure}[t!]
    \centering
    \includegraphics[width=\columnwidth]{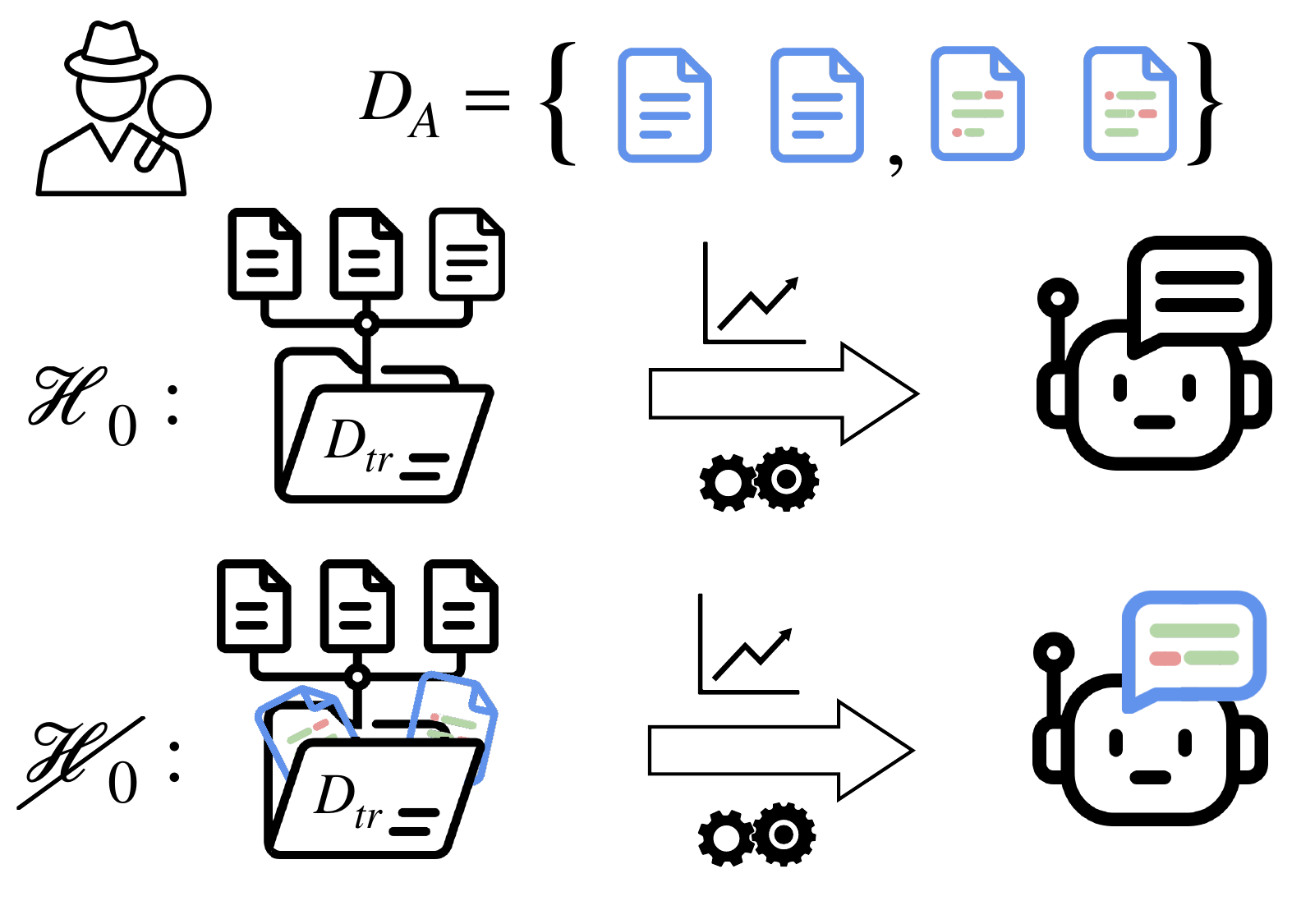}
    \caption{To protect a proprietary dataset from unauthorized use in training, the dataset owner (attacker) paraphrases their documents with a secret watermark key. To perform dataset inference, the attacker tests the suspect model's predictions for evidence of the watermark key. The watermark detection test is used to conclude whether their protected data was included in the training dataset.}\label{fig:method-explainer}
\end{figure}

\textbf{What makes training dataset membership tests challenging in modern settings?}
Performing membership tests on modern generative models is harder than it was in discriminative settings for classifiers which mapped inputs to as few as 10 or 100 classes (i.e. the setting of \citet{shokri2017membership}). The variable-length output space matches the input cardinality, making analysis complicated, and the billion-parameter sizes of modern models make classical attribution tools like influence functions difficult to apply \citep{koh2017understanding,ilyas2022datamodels,park2023trak}. Even the definition of ``a sample'' is arbitrary at pretraining scale (documents, sentences, words), and individual samples overlap heavily in terms of shared n-grams, which destabilizes existing membership tests \citep{duan2024membership}. Modern generative models also memorize \textit{and} generalize along both semantic and stylistic lines, blurring the line between membership and generation attributability: sample membership is often neither necessary nor sufficient for a target output to be produced \citep{liu2025language}, and memorization rates vary widely \citep{cooper2025extracting}.

\textbf{Proactive Interventions for Reliable Dataset Inference.}
Our work attempts to circumvent some of the difficulties described above by proactively but sparingly intervening during the construction of training datasets to make membership tests easier. The essential operation will be to precisely \textit{mark} the training dataset of a model in a way that does not significantly impact performance. This intervention will give the marker (data owner) a significant informational advantage about how the marked elements of the training dataset are expected to influence the model's test-time predictions, and most importantly, how to probe for these effects. 

The main technical gadget we will use is language model output watermarking of the kind proposed and studied in \citet{kirchenbauer23a}. This watermarked decoding algorithm will be layered on top of a ``paraphraser'' language model to create versions of candidate samples with the same style and semantics, but that produce significant p-values under a watermark detection test. Recent work has shown that if a target model is trained on watermarked samples, it will absorb the watermark key's signature and emit a shadow of the same signature at test time \citep{sander2024watermarking}. Then, when the attacker probes a suspect model using the watermark detection test, sample membership in this model's training dataset is inferrable based on the $p$-value observed.

\paragraph{Relationship to Prior Work.}\label{sec:rel-to-prior-work}
We build our study directly on two prior works. The folding construction used as our finetuning backbone is adapted from \citet{hayes2025exploring}, whose paired-data MIA evaluation we extend to full-subset membership (DIA). The proactive watermarking DI approach builds on \citet{sander2024watermarking,sander2025detecting}, who show that a target model trained on watermarked samples measurably reproduces the watermark's key-specific signature. We adopt their sensitive ``reading-mode'' detector that tests next-token predictions (argmaxes) conditioned on watermarked prefixes rather than fully rolled-out completions. We vary per-key support fractions and effective epochs on the marked subsets, replace the detection test's parametric tail with an empirical-null randomization test (\cref{sec:p-value-validity}), and benchmark against loss-based and reference-model baselines---raw loss \citep{yeom2018privacy}, argmax match, min-$k\%$ \citep{shi2024detecting}, zlib \citep{carlini2021extracting}, and rMIA-simple, rMIA, and LiRA \citep{carlini2022membership}---on the same paired trials (\cref{sec:appendix-phase1-event-baselines}). \textbf{Our goal is to unify these settings and methods to calibrate claims about whether watermark-based dataset membership testing is ready for use in practice.}

\textbf{Contributions:}
\begin{itemize}[topsep=0.0cm,itemsep=-0.1cm,leftmargin=0.4cm,font=\bfseries]
\item We formalize the general threat model for watermark-based proactive dataset inference as a data-owner-centric security game, and contextualize prior work as studying a specific instance of this problem (\cref{sec:method}).
\item We propose a randomization-based variant of the watermark detection test that ensures $p$-value validity under interactions between pretrained checkpoints, intervention samples, and specific watermark keys (\cref{sec:p-value-validity}).
\item We implement a unified experimental setting for evaluating traditional loss-based dataset inference scores and watermark-based approaches on the same paired trials, and compare the performance of each technique under its respective threat-model assumptions (\cref{sec:phase1-event-baselines}).
\item We ablate under-explored dimensions from previous studies: per-key support vs. repetition at fixed training budget, training from scratch vs. continued pretraining, and insertion schedule shape for the marked subset (\cref{sec:phase1,sec:phase2}).
\end{itemize}

\section{Methodology}\label{sec:method}

\begin{definition}[\textbf{\textit{Threat Model}: Proactive Dataset Inference for Data Owners.}]
Alice suspects that a model owner Bob will train a language model $f_{tgt}$ on $D_{tr} := D_{A} \cup D_{web}$, where $D_A$ is a portion Alice owns or controls and the rest is drawn from sources Alice does not control. Alice wants a score $d: f_{tgt},\mathcal{D}_{A} \to (0,1)$ that predicts whether Bob included any of $D_{A}$ in $D_{tr}$. Alice is allowed to modify $D_A$ using a \textit{marking} transformation $T_m: \mathcal{X} \to \mathcal{X}_m$ before Bob accesses it. Attack success is evaluated against the ground-truth membership label $m \in \{0,1\}$ indicating $D_{A}$'s inclusion in $D_{tr}$. 
In the strictest setting, Alice has neither the compute to train shadow models for $f_{tgt}$ nor a hidden shadow copy of similar data $D'_{A}$ to calibrate the detector against.
\end{definition}

\begin{definition}[\textbf{\textit{Gadget}: Watermark-based Dataset Protection.}]\label{def:gadget}
Alice instantiates $T_m$ using a generative model $f_{wm}: \mathcal{X},k \to \mathcal{X}_{wm}$ running an output watermarking scheme keyed by secret key $k$, whose detector $d(f_{tgt}, k,D_{A})$ admits a test statistic $z \in \mathbb{R}$ and associated $p$-value. She marks all or part of $D_{A}$ via $f_{wm}$ relying on the assumption that Bob's $f_{tgt}$ trained on samples $x \in \mathcal{X}_{wm}$ will produce new generations that yield significant detection scores under $k$. A decision function $g: \mathbb{R} \to (0,1)$ on $z$ then implements the membership score $d = g(z)$, with the $p$-value itself being the obvious choice. The null hypothesis is defined as $\mathcal{H}_0:$ Bob's model $f_{tgt}$ was trained such that $D_A \cap D_{tr} = \emptyset$, or more precisely, such that none of the samples in $D_{tr}$ are marked with $k$. Assuming the canonical choice of $g(\cdot)$, if a $p$-value below $\alpha$ is observed, Alice rejects the null and concludes that Bob likely trained $f_{tgt}$ on her dataset.
\end{definition}

\subsection{Target Model Assumptions}\label{sec:target-model-assumptions}

We assume that the target model's tokenizer is known and matches the paraphraser's. Prior work already shows that this assumption can be relaxed and watermark detectability still preserved through common-token filtering \citep{sander2024watermarking,sander2025detecting,xu2026antidistillation}, so we omit experiments along this axis here. We also assume that the attacker can query the target model for next-token-prediction probabilities conditioned on a prefix---the ``reading mode'' detection procedure of \citet{sander2024watermarking}. This is a costly assumption in practice if one is making calls to a production API for every token. However, the increased sensitivity of the data-forced reading-mode test is necessary at the support fractions we consider: naively rolled-out completions from a target model that has only been mildly exposed to the watermark signature do not yield low enough $p$-values for reliable detection. Improvements in detection that would let the attacker relax the reading-mode assumption are an obvious axis of future work but are out of scope here.

\subsection{Ensuring P-value Validity}\label{sec:p-value-validity}

Prior work on watermark-based DIAs reports significant $p$-values under intervention and near-null values otherwise \citep{sander2025detecting}. However, our preliminary reproduction experiments suggested that interactions between the model checkpoint, the particular samples in the intervention dataset, and the specific watermark key used can invalidate the independence assumptions on which standard parametric tail bounds depend, even under careful de-duplication and especially when the reading-mode detector is in use. Therefore, we propose a standard randomization-based (permutation) hypothesis test variant that establishes a valid $p$-value based on an empirical null hypothesis under exchangeability assumptions. This method allows us to configure the resolution of the exact $p$-value and adds only a small computational overhead to the attacker's testing protocol.

\begin{definition}[\textbf{Watermark Key Randomization Test for Exact P-Values}]
Assume Alice's watermark secret $k_i$ is sampled i.i.d. from a distribution of possible keys suitable for the PRF used by the watermark scheme $k_i \sim \mathcal{K}$. Under the null hypothesis, $f_{tgt}$ has not been trained on data marked with $k_i$ nor any other key $k_j$ in $\mathcal{K}$\footnote{This is a strong, but plausible assumption if a sufficiently large, private keyspace is used.}. Thus all keys $k_j \neq k_i$ are exchangeable w.r.t. the test statistic produced by the detector $d(f_{tgt},k,D_{A})$. To compute the exact p-value using the randomization test, Alice samples $M-1$ distinct additional keys, and then produces $z_j = d(f_{tgt},k_j,D_{A}), \forall j \in M$ including her key $k_i$. The $M$ scores are sorted in descending order to determine each key's resulting rank $r_j$ (highest $z$ yields smallest $r$). The exact p-value of the test is $r_i/M$, the rank of Alice's actual key amongst $M$. Under the null, due to exchangeability, the probability of observing her key at rank $r_i$ or earlier in the list is exactly $P[r \leq r_i | \mathcal{H}_0] = r_i/M$, making this a valid p-value under no additional assumptions.
\end{definition}

Test-time cost is controlled using the vectorized PRF implementation in \texttt{textseal} \citep{sander2025detecting,fernandez2025how} to evaluate many watermark secrets in parallel for any given piece of text. We verify the validity of the resulting empirical null in practice by confirming $p$-value uniformity using a KS test (see \cref{fig:event-null-validity,fig:appendix-event-null-validity,fig:sks-null-validity}). For $p$-values below the $1/M$ resolution of the empirical sample, we report an extrapolated empirical-Gaussian fit to the same null distribution where useful.

\begin{figure*}[ht!]
    \centering
    \includegraphics[width=0.49\textwidth]{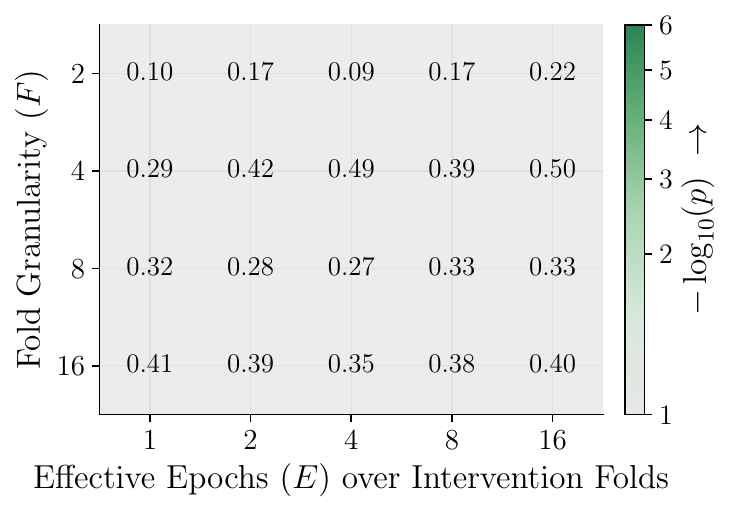}
    \hfill
    \includegraphics[width=0.49\textwidth]{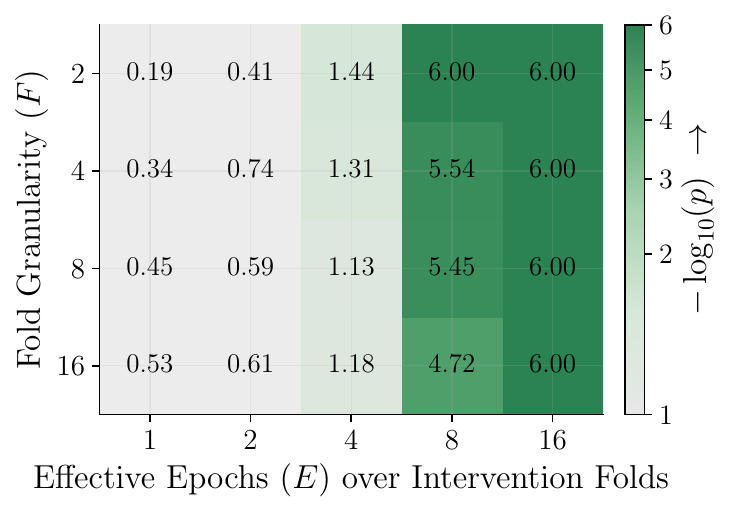}
    \caption{Finetuning event-split matched clean-twin false-probe null \textbf{(left)} and keyed signal \textbf{(right)} on the aligned unpacked detection surface, scored as $-\log_{10} p$ under an empirical-exact reference.}
    \label{fig:event-keyed-aligned-exact}
\end{figure*}

\section{Experiments}\label{sec:experiments}

Our experiments are built around a controlled data-folding design that allows us to mimic prior studies on membership and dataset inference in language models while simultaneously benchmarking the proactive watermarking approach. In a smaller finetuning regime, we systematically vary the level at which the model is exposed to the intervention data by modulating subset size and repetition during training, before moving to a larger 10B-token training regime where we ablate the insertion schedule of the marked subset and the choice of initialization (continued pretraining versus from-scratch).

\paragraph{Setup.}
In our experiments, we strive for depth rather than breadth. Thus, we use a single language model \texttt{Qwen3/Qwen3-1.7B} \citep{yang2025qwen3} in all experiments, either as a pretrained set of weights or as the architectural specification for from-scratch experiments. Similarly, we adopt a single dataset of $N=1500$ natural-looking but semantically isolated documents designed to be used in controlled experiments on memorization: FictionalQA \citep{kirchenbauer2025fictional}. This dataset includes webtext-like documents in generic styles about totally fictional entities and \textbf{events}---collections of related documents about the same fictional scenario (an explanatory blog post, a news article about the event, etc.). Our \emph{event-split} fold construction (\cref{sec:phase1}) uses the event groupings present in the dataset: fold boundaries respect events so that documents about the same fictional event stay together inside a single fold. We mix our controlled intervention data into a base random subset of \texttt{allenai/dolma3\_mix-150B-1025} \citep{olmo2025olmo3} with both sources shuffled and packed into length-4096 chunks.

\paragraph{Dataset Membership Testing.}
During each dataset inference simulation trial, for the watermarking methods, for positive cases we report the pooled, deduplicated reading-mode test statistic and corresponding $p$-value measured on each of the watermarked folds for each of the models that trained on it. For the negative cases, we compute the same score using all the same watermarked folds as probe data, but on the corresponding clean (twin) model trained on the unmarked version of the data. For the non-watermarking methods, the original clean samples from the fold are fed through the corresponding clean models to compute losses, and then each technique's corresponding algorithm is used to produce the row-level MIA or whole-model DIA score, potentially using the rest of the clean models as reference models. Throughout, we present results in terms of $-\log_{10} p$ for keyed signal strength, reporting the exact empirical-null reference distribution (\cref{sec:p-value-validity}) and/or an extrapolated empirical-Gaussian p-value where useful.

\begin{figure*}[h!]
    \centering
    \includegraphics[width=0.49\textwidth]{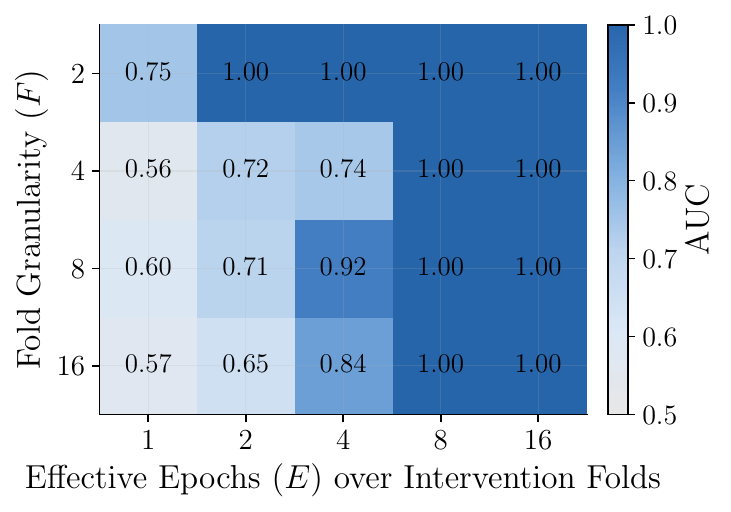}
    \hfill
    \includegraphics[width=0.49\textwidth]{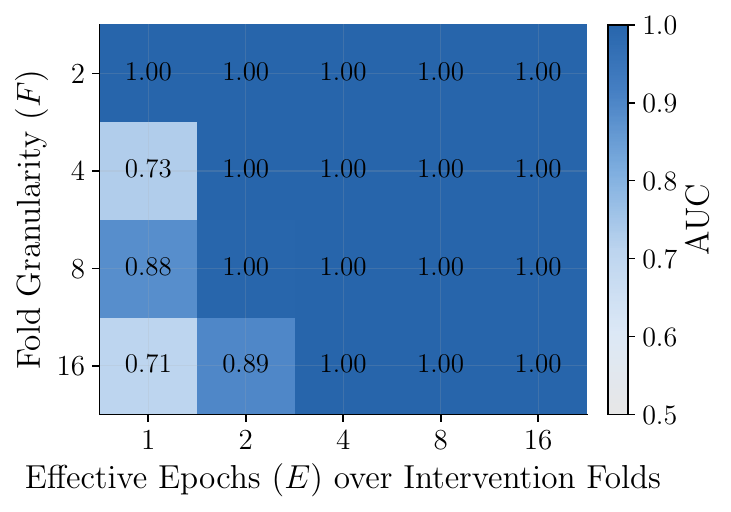}
    \caption{Finetuning event-split watermark whole-model DIA AUC across the $F \times E$ grid, on the aligned \textbf{(left)} and packed \textbf{(right)} detection surfaces, both scored against an empirical-exact null. The aligned chart on the left is based on the same probe surface as the right panel of \cref{fig:event-keyed-aligned-exact}, paired here with the packed-surface companion. On the aligned surface the AUC saturates in the high-exposure corner; on the packed surface the more permissive oracle recovers several lower-exposure cells.}
    \label{fig:event-dia-exact}
\end{figure*}

We score the watermark detection statistic under two surface variants. The \textbf{aligned} surface scores each original document in isolation and is the most realistic setting; Alice has no knowledge of exactly how her documents will be truncated or concatenated together. The \textbf{packed} surface scores the 4096-token chunks of multiple fictional documents packed exactly as they were at training time, and serves as an oracle baseline that quantifies how much keyed signal could be lost to train- vs.\ inference-time mismatches. We score the same readout under both an empirical-Gaussian and an exact empirical-null reference distribution (\cref{sec:p-value-validity}), and report headline results in $-\log_{10} p$ throughout (higher means a more significant signal).
\begin{table}[hb!]
\centering
\caption{Idealized per-key exposure $E/F$ for the finetuning grid, expressed as a multiple of the per-key support that a single epoch over a single fold would produce. Distinct watermarked tokens scale as $1/F$; epoch repetition $E$ multiplies how many times those tokens are seen during training. The event-split and SKS regimes share this idealized table. Realized exposure based on the actual sampling rates during the training runs is reported in \cref{sec:appendix-phase1-event-ehat,sec:appendix-phase1-sks-ehat}.}
\label{tab:phase1-sks-idealized-exposure}
\begin{tabular}{c|ccccc}
 & $E=1$ & $E=2$ & $E=4$ & $E=8$ & $E=16$ \\
\hline
$F=2$  & 0.5    & 1.0   & 2.0   & 4.0   & 8.0 \\
$F=4$  & 0.25   & 0.5   & 1.0   & 2.0   & 4.0 \\
$F=8$  & 0.125  & 0.25  & 0.5   & 1.0   & 2.0 \\
$F=16$ & 0.0625 & 0.125 & 0.25  & 0.5   & 1.0 \\
\end{tabular}
\end{table}

\subsection{Finetuning: Varying Per-Key Exposure}\label{sec:phase1}

In our finetuning-scale event-split design, multiple keyed folds of size $N/F$ are mixed into each model's training data with the total watermarked share of the training tokens held fixed across $F$. As $F$ grows, each individual key's per-fold footprint shrinks while the number of distinct keys observed by the model grows. We populate every cell of the $F \times E$ grid with enough watermarked-positive and matched clean-false-probe trials to make whole-model dataset inference comparisons stable. The per-cell training token budget is held fixed at 131M tokens. The interventional design is summarized in \cref{tab:phase1-sks-idealized-exposure} (idealized per-key exposure $E/F$, shared with the SKS ablation); the matched realized normalized exposure $\hat{E}/F$ and the underlying realized $\hat{E}$ values live in the appendix (\cref{tab:wm4memb-phase1-event-grid-fixed-exposure-current,tab:wm4memb-phase1-event-grid-fixed-ehat-current}). In absolute terms, the watermarked share of each model's training tokens swings from a low of about 71k tokens (${\approx}0.05\%$ of the run) at $(F{=}16, E{=}1)$ up to about 8.3M tokens (${\approx}6.3\%$) at $(F{=}2, E{=}16)$, and per-cell whole-model DIA trial counts scale with $F$, from $2 / 2$ positive/negative trials per cell at $F{=}2$ to $96 / 96$ at $F{=}16$ (\cref{tab:wm4memb-phase1-event-fixed-training-scale-context,tab:wm4memb-phase1-event-fixed-dia-context}).

\paragraph{Increased signal as a function of exposure ($E/F$).}
\Cref{fig:event-keyed-aligned-exact} demonstrates the expected trends: increasing $E$ produces strongly significant keyed signal whenever the underlying support is sufficient, while increasing $F$ at fixed $E$ weakens the signal as each individual key occupies less of the corpus (\cref{fig:event-trend} makes this more obvious). The DIA AUC view (\cref{fig:event-dia-exact}) tracks the keyed-signal ordering and saturates at $1.0$ in the high-exposure corner, while the low-exposure corner sits near chance. The matching whole-model DIA comparison against loss-based and reference-model baselines on the same paired trials is reported in \cref{tab:wm4memb-phase1-event-grid-fixed-dia-current}; the corresponding row-level MIA, matched false-probe-null, cross-key sham-null, and packed-surface counterparts are deferred to the appendix in \cref{sec:appendix-phase1-event}. 
The single-key support (SKS) ablation, which holds each model to one watermarked fold so per-model fictional support drops as $1/F$, is reported in \cref{sec:appendix-phase1-sks} as a structurally simpler reference point.

\paragraph{Event-Split Null-Validity}\label{sec:phase1-event-null-validity}

\begin{figure}[ht!]
    \centering
    \includegraphics[width=0.9\columnwidth, trim=0cm 0cm 10.5cm 0cm, clip]{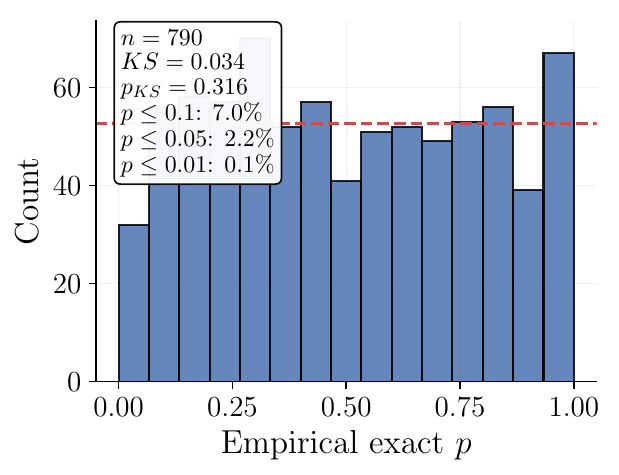}
    \caption{Event-grid null-validity panel. Histogram of pooled empirical exact $p$-values from the packed matched-clean-negative whole-model readings across the $(F,E)$ event grid. The annotated statistic and $p$-value are from a one-sample Kolmogorov-Smirnov test of these pooled $p$-values against $\mathrm{Uniform}(0, 1)$; the dashed horizontal line is the expected per-bin count under a uniform histogram with the plotted binning.}
    \label{fig:event-null-validity}
\end{figure}

\Cref{fig:event-null-validity} confirms that, once pooled across many distinct positive keys, the empirical exact null is close to uniform at the 1M-null scale and respects standard tail-rate thresholds, supporting the use of the empirical-null permutation test as the headline reference distribution for the keyed readout. Per-key idiosyncrasy is a separate concern visible in the right-hand trace of the extended chart \cref{fig:appendix-event-null-validity}: at very small fold counts a single key can run systematically warm even when the pooled null is well-calibrated, and the inherited $F=2$ scaffold reused in pretraining (\cref{sec:phase2}) carries one such warm key.

\paragraph{Loss-Based and Reference-Model Baselines}\label{sec:phase1-event-baselines}

\Cref{tab:wm4memb-phase1-event-grid-fixed-dia-current} reports the fold-level whole-model DIA comparisons for the event-split finetuning grid (for the decision problem ``was model $i$ trained on fold $i$?''), with the watermark method reported alongside the loss-based and reference-model baselines on the same paired trials. The watermark whole-model DIA AUC values reproduce those visualized in \cref{fig:event-dia-exact}. Row-level MIA performance for the loss-based methods is reported in \Cref{tab:wm4memb-phase1-event-grid-fixed-row-baselines-current}. Using additional access to calibration samples (``Loss-based'' ref-free) and/or the rest of the models in the pool trained and not trained on each fold as a reference pool (``Ref-model''), the baselines perform nearly perfectly with the exception of the very low-support $F8$ and $F16$ single-epoch cells. While valid to run above $F=2$, LiRA still suffers in the $F4$ cells when the ref-model pool is too small for accurate in/out ratio estimation (see \cref{tab:wm4memb-phase1-event-fixed-dia-context}). 
We also render a compact summary table of access assumptions for the various approaches in the appendix (\cref{tab:access-assumptions}).

\input{auto_figures/wm4memb_phase1_event_grid_fixed_dia_current.tex}

\subsection{Pretraining: Sensitivity at Scale}\label{sec:phase2}

To address whether the keyed readout remains detectable under much heavier dilution, we conduct a second, narrower batch of experiments at a 10B-token total budget. We reuse a similar $F=2$ fold/key scaffold from finetuning (two groups of randomly sampled fictional documents of size $N/2$, simpler than the event-split construction) and sweep ten different insertion schedules for these two keyed folds.

The two initialization regimes are continued pretraining (CPT) from the Qwen3-1.7B checkpoint and from-scratch (random init) at the same total token budget.
The four single-burst schedules $(S1,E1)$, $(S2,E1)$, $(S3,E1)$, and $(S4,E1)$ insert the watermarked fold once at one of four step locations within the run (target $E = 1$). The remaining six schedules pair a uniform-spacing variant (denoted $U$), which uses a constant sampling rate targeting $\sim E$ epochs by end of run, with a periodic-cluster variant (denoted $P$) that schedules $E$ dumps of the entire fold at evenly spaced steps, very similar to \citet{sander2025detecting}'s setup. We cross these with $E \in \{4, 8, 16\}$. The idealized exposure profile of these schedules at $F=2$ is summarized in \cref{tab:phase2-idealized-exposure}; the matched realized normalized exposure tables and per-schedule realized $\hat{E}$ values for both initialization regimes also live in the appendix (\cref{tab:wm4memb-phase2-cpt-exposure,tab:wm4memb-phase2-scratch-exposure,tab:wm4memb-phase2-cpt-ehat,tab:wm4memb-phase2-scratch-ehat}). 

Main body figures use the exact empirical $p$-value for both CPT and from-scratch training. Each schedule contributes 2 watermarked-positive and 2 matched clean-negative whole-model trials per init, and (mirroring \citet{sander2025detecting}) the watermarked-token share of each 10.49B-token training run is small in absolute terms---about $0.5M$ tokens (${\approx}0.005\%$) for the four single-burst $E{=}1$ schedules, rising to about $8M$ tokens (${\approx}0.077\%$) at the $E{=}16$ schedules (\cref{tab:wm4memb-phase2-cpt-training-scale-context,tab:wm4memb-phase2-cpt-dia-context,tab:wm4memb-phase2-scratch-training-scale-context,tab:wm4memb-phase2-scratch-dia-context}).

\begin{figure*}[ht!]
    \centering
    \includegraphics[width=0.49\textwidth]{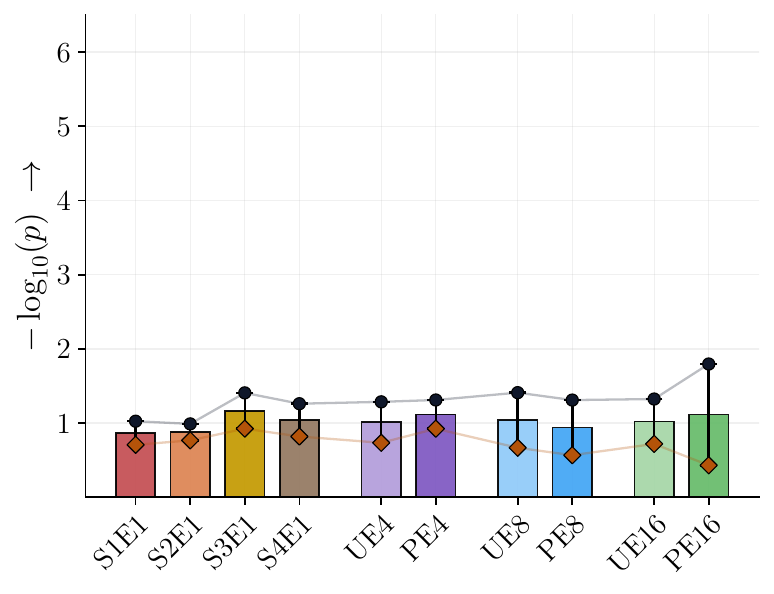}
    \hfill
    \includegraphics[width=0.49\textwidth]{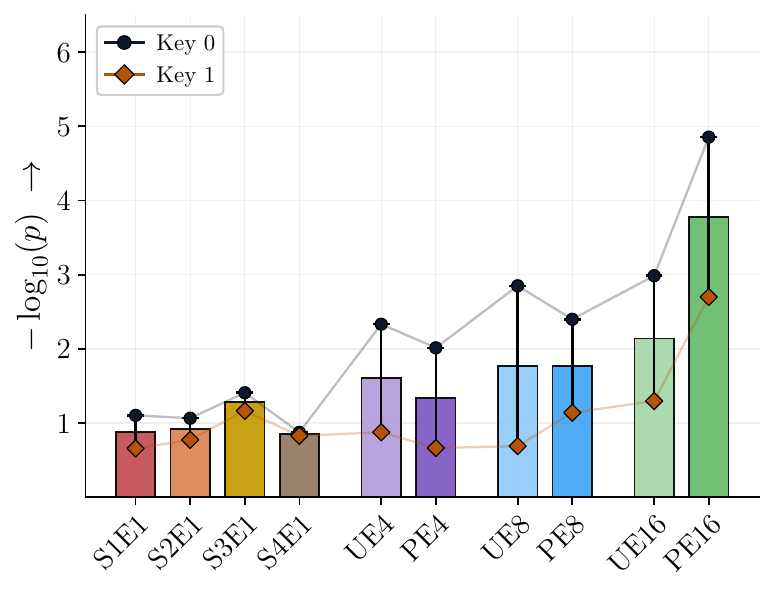}
    \caption{Pretraining signal across the ten-schedule sweep on the aligned unpacked detection surface, scored as $-\log_{10} p$ under the exact empirical-null reference, for the CPT initialization. \textbf{(Left)} is the clean-twin false-probe-null models and \textbf{(right)} is the keyed model trained on the watermarked data. Schedule clearly modulates the keyed readout: at $E1$, none of the keyed signals are separably significant, but at larger $E$ a gap appears. One of the two inherited keys runs visibly warmer than the other in both panels. Matched packed-surface variants, extrapolated Gaussian p-value versions, and watermark whole-model DIA bar charts are all reported in the appendix (\cref{sec:appendix-phase2-cpt-companions})}
    \label{fig:phase2-cpt-aligned-exact}
\end{figure*}

\begin{figure*}[ht!]
    \centering
    \includegraphics[width=0.49\textwidth]{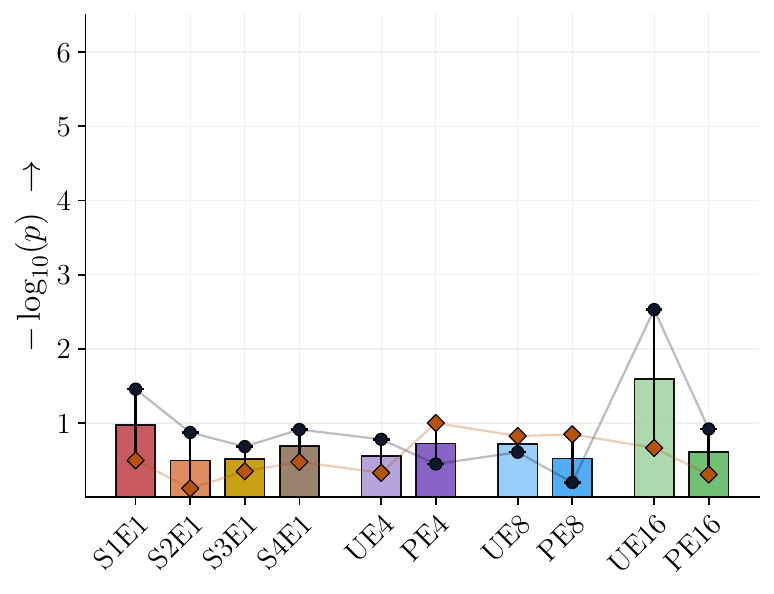}
    \hfill
    \includegraphics[width=0.49\textwidth]{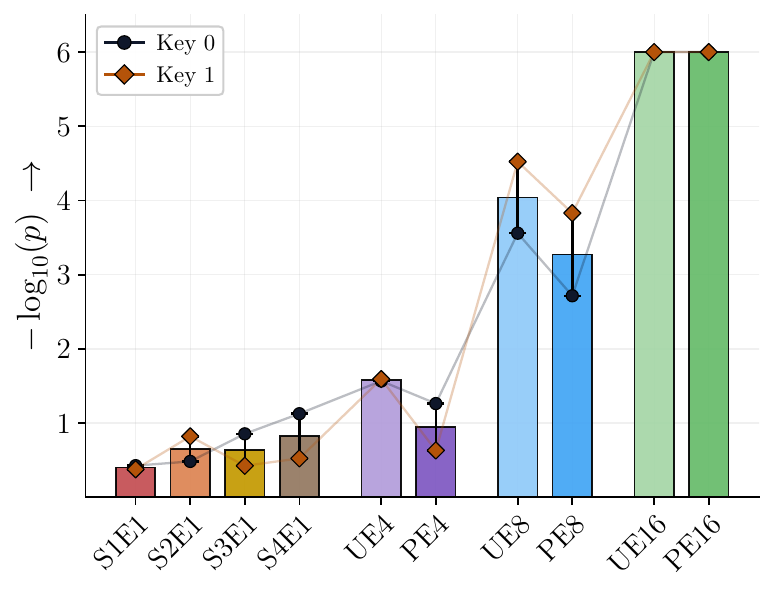}
    \caption{Pretraining signal across the ten-schedule sweep on the aligned unpacked detection surface, scored as $-\log_{10} p$ under the exact empirical-null reference, for the from-scratch initialization. \textbf{(Left)} is the clean-twin false-probe-null models and \textbf{(right)} is the keyed model trained on the watermarked data. The schedule strengthens the keyed readout much faster as a function of $E$ in this setting versus CPT: at $E1$ there is a slight trend as a function of insertion point (\cref{fig:appendix-phase2-scratch-packed-exact} makes this more obvious) at $E8$ a significant gap appears and $E16$ maxes out the exact resolution. Here, there is little if any systematic bias between the watermark keys. Matched packed-surface variants, extrapolated Gaussian p-value versions, and watermark whole-model DIA bar charts are all reported in the appendix (\cref{sec:appendix-phase2-scratch-companions})}
    \label{fig:phase2-scratch-aligned-exact}
\end{figure*}

\paragraph{Schedule and Initialization Effects.}
Across both inits, schedule shape matters: the four single-burst $E{=}1$ schedules are not distinguishably higher than their null counterparts in \cref{fig:phase2-cpt-aligned-exact}, while the multi-insertion $U$ and $P$ schedules become clearly separable from null as $\hat E$ grows. Under aligned testing the $P$ and $U$ schedules are similar, but the more sensitive packed-surface readings (\cref{fig:appendix-phase2-cpt-packed-exact,fig:appendix-phase2-cpt-packed-gaussian}) show that the $P$ setting produces a stronger signal for CPT. From-scratch recovers substantially stronger keyed signal than CPT at high $\hat E$, while at low $\hat E$ the two inits are comparable, though \cref{fig:phase2-scratch-aligned-exact} shows slightly more separability at $\hat E=4$ for from-scratch than for CPT. \Cref{fig:appendix-phase2-scratch-packed-exact,fig:appendix-phase2-scratch-packed-gaussian} show that the $U$ sampling strategy produces more signal in the from-scratch setting, the opposite of the CPT result. 

\paragraph{Fold/Key Asymmetry.}
The pretraining experiments reuse the same two folds and keys that the finetuning SKS $F=2$ row was conducted with. At the finetuning scale, one of those two keys already produces systematically larger detection statistics than the other under matched-clean conditions, even when no watermark training signal is in play (right side of \cref{fig:appendix-event-null-validity,fig:sks-null-validity}). This handedness is visible again in \cref{fig:phase2-cpt-aligned-exact}, where Key 0 consistently produces higher signal than the other under matched conditions (more obvious in \cref{fig:appendix-phase2-cpt-packed-exact}). This suggests that per-key interactions with the data and/or the pretrained checkpoint can produce scenarios where positive model signal is spuriously low or high, or false-positive rates could exceed what the nominal $p$-value threshold would indicate. It is possible that prior studies filtered for well-behaved watermark keys, and our systematic evaluation just reports this issue more transparently.

\paragraph{Comparison to loss-based baselines.} A complete loss-based and reference-model whole-model DIA head-to-head comparison for both pretraining inits is reported in the appendix (\cref{sec:appendix-phase2-dia,tab:wm4memb-phase2-cpt-dia-baselines,tab:wm4memb-phase2-scratch-dia-baselines}). As in the finetuning regime, the loss-based approaches perform nearly perfectly across all pretraining schedules including the low-exposure $E1$ insertions. As the differences between \cref{fig:phase2-scratch-aligned-exact} and \cref{fig:phase2-cpt-aligned-exact} would suggest, the watermark-based approach is only able to separate the positives and negatives clearly when the models are trained from scratch at the higher-exposure $E8$ and $E16$ settings.

\section{Discussion}

As stated in the introduction, our setup unifies and builds upon two prior studies: \citet{sander2025detecting} and \citet{hayes2025exploring}, rather than proposing a wholly novel methodology. Therefore, it is important to identify where our results cohere with and diverge from those studies. After presenting our analysis, we discuss the feasibility of Bob, the model trainer, ``defending'' against this dataset protection strategy.


\paragraph{Watermark Detectability Trends.}
The trends we observe in both \cref{sec:phase1} and \cref{sec:phase2} on sensitivity growing with exposure to the watermarked subset match expectations from \citet{sander2025detecting}. Even though that work does not directly ablate the size of the watermarked subset as we do at finetuning scales, we expect that, were resources sufficient to test it, the drop in significance as a function of increasing $F$ that we observe in \cref{sec:phase1} would also appear at the pretraining scale. 

\paragraph{Performance of Loss-based Methods.}
\looseness -1 While~\citet{hayes2025exploring} did not run as complete a controlled exposure grid as our $(F,E)$ sweep in \cref{sec:phase1}, the loss-based methods' row-level MIA success across the exposure grid varies in the very same way as the watermark signal strength (\cref{sec:appendix-phase1-event-baselines}). However, they almost perfectly solve the DI problem at all exposure levels anyway (\cref{tab:wm4memb-phase1-event-grid-fixed-dia-current}). As noted in \citet{hayes2025exploring}, LiRA struggles at low ref-model counts.

\paragraph{Sensitivity at Scale and FPR Control Issues.}
We observe relatively similar $-\log_{10} p$ values for $PE4$, $PE8$, and $PE16$ for the models trained on the watermarked data in \cref{sec:phase2} to analogous values reported in \citet{sander2025detecting} (our fold size vs. training budget are approximately matched here). However, this is only true when the model is trained from scratch, something \citet{sander2025detecting} did not ablate, and significance swings sharply depending on train- vs.\ detect-probe style (aligned/packed). The finetuning and CPT settings also reveal that a spuriously hot watermark key might confound the results or cause false positives (see \cref{sec:appendix-phase1-event-null-validity,sec:appendix-phase1-sks-null-validity}). The point estimates reported in \citet{sander2025detecting} make it hard to determine whether this was an issue in their study. However, they do report a $-\log_{10} p$ range of $0.3$ to $0.9$ in Table 1 at zero contaminations which is broadly in line with our \cref{fig:phase2-scratch-aligned-exact} null side readings, but crucially this is in the from-scratch setting where there is no preexisting pretrained checkpoint for the specific watermark key to potentially interact with. Our results show that even under a well-calibrated exact p-value test, spurious key-checkpoint interactions still matter for any single key in any single deployment attempt. Therefore, Alice may need to filter for ``quiet'' keys that behave well under the null hypothesis for representative data and models.

\paragraph{Can the Model Trainer Mount a Defense?}

The marking transformation $T_m: \mathcal{X} \to \mathcal{X}_m$ is intentionally notated to bring attention to the fact that $\mathcal{X}$ and $\mathcal{X}_m$ are distinguishable, \textit{but only to Alice}. 
While a data owner does not necessarily need to mark their data in a way that satisfies a formal third-party indistinguishability condition (a la \citet{christ2024undetectable}) for the attack to be successful, in practice, a marking scheme where the model trainer is unable to reliably (or \textit{scalably}) determine whether or not a sample is in $\mathcal{X}_m$ is likely to be much more effective and tamper-proof. If Bob cannot tell the difference between samples in $D_{tr}$ that are marked or not, then Bob has no way of knowing whether or not Alice has marked the data, nor can Bob efficiently filter the dataset to remove $D_A$ from it.

If we assume that there exist watermarking methods such that i) watermarked outputs have the same quality as non-watermarked samples, ii) watermarked samples are indistinguishable from non-watermarked samples for computationally-bounded discriminators, and such that iii) an observer can readily tell the distributions apart if the observer does possess the watermark secret, then a paraphrasing model equipped with an output watermark may be \textit{the} theoretically optimal marking transformation $T_m$ (see \citet{christ2024undetectable,piet2025markmywords} for some discussion of these ideas in more context). 
Such an invisible, ``non-distortionary'' watermark family might serve as a better instantiation of $T_m$ in the proactive dataset protection methodology we study in this work. However, whether the the implicit ``learnability'' required for the marking to be detectable in Bob's trained model is in fundamental tension with indistinguishability properties is an open question and a fruitful topic for future research.

\section{Conclusion}

We find the watermark-based dataset inference approach to be a promising alternative to traditional loss-based DIA methods, but overlooked details in previous studies highlight important avenues for future research. Despite our randomization-based detection test yielding a well-calibrated empirical null, the possibility for elevated single-key readings in any setting presents a previously overlooked deployment challenge. Further, any train- vs.\ detection-time mismatch in document presentation (aligned/packed) can reduce test significance values. Finally, the reading-mode detection procedure from prior work that is required for signal at high dilution may itself be impractical depending on the API access the data owner has against the suspect target model. 

In head-to-head comparisons under a controlled experimental design, the loss-based baselines are more performant as DIA methods under an ROC-AUC evaluation. However, by definition they do not admit p-values for assumption-free and calibration-free FPR estimation and they assume greater access to the target model (loss values), some known non-member, distributionally-matched calibration samples, and/or a pool of trained reference models (\cref{tab:access-assumptions}). Therefore, which approach is ultimately more useful in practice comes down to which set of assumptions is more realistic for the particular data owner and what level of sensitivity under dilution their use-case requires.


\section*{Acknowledgments}

This work was supported by DARPA TIAMAT, the NSF TRAILS Institute (2229885), and Coefficient Giving. Computing resources for this project were supported in part by the U.S. Department of Energy, Office of Science, Office of Advanced Scientific Computing Research under the Advancements in AI for Science program.

Prepared by LLNL under Contract DE-AC52-07NA27344 and supported by the LLNL-LDRD Program under Projects No. 24-ERD-010, and 24-ERD-058 (LLNL-CONF-2019479). This manuscript has been authored by Lawrence Livermore National Security, LLC under Contract No. DE-AC52-07NA27344 with the U.S. Department of Energy. The United States Government retains, and the publisher, by accepting the article for publication, acknowledges that the United States Government retains a non-exclusive, paid-up, irrevocable, world-wide license to publish or reproduce the published form of this manuscript, or allow others to do so, for United States Government purposes.

\textbf{AI Use}: Models from both OpenAI (Codex) and Anthropic (Claude) were used extensively in the implementation, experiment management, and paper editing aspects of this work.

\bibliography{references}
\bibliographystyle{icml2026}

\newpage
\appendix
\onecolumn

\section{Extended Methodological Details}\label{app:ext-method-details}

\paragraph{Folded Data Design.}\label{app:data-fold-design}

We instantiate our folded design under two complementary support regimes that share the same $(F, E)$ axes. The primary, more realistic regime is the \textit{event-split} setting: multiple keyed folds are mixed into each model's training data with event-level groupings preserved within each fold and with the total watermarked share of the training tokens held fixed across $F$. As $F$ grows, each individual key's per-fold footprint shrinks while the number of distinct keys present grows, simulating a deployment scenario where multiple data owners protect their own subsets with distinct keys inside a single training corpus. The simpler ablation regime is the \textit{simple per-key support} (SKS) setting: each watermarked model trains on exactly one watermarked fold, which isolates the watermarked-fraction-versus-repetition question without sibling-key interference but is unrealistic in that no real deployed corpus would protect only one fold of one data owner's content under a single key. We treat the SKS setting as a structurally clean warm-up that anchors the more realistic event-split readout.

\paragraph{Efficiency-Focused Design Choices.}
The point of this folded design is to realize a large number of dataset inference trials simulating the proactive watermarking approach as well as running the loss-based methods, all while still controlling computational cost. As a result, certain independence assumptions are necessarily violated. First, the raw source data used at each $F$ level is the same fixed pool of FictionalQA documents, which makes the different experiments correlated with respect to this domain. Second, in the event-split regime each model trains on more than one fold, so a dataset inference experiment for that model with respect to fold $i$ is not fully independent of the experiment testing the same model with respect to fold $j$ which it was also trained on.

However, some of the efficiency-focused choices also help control other confounds. As we increase $E$, the fold partitions (actual selected documents) and watermark keys used remain identical, making effective epochs the only variable along that axis. Similarly, the folds that the paired clean models train on are the same underlying documents as the ones that the watermarked models train on (just before they were watermarked), controlling for natural variation in modeling difficulty between the watermarked and clean model pairs. In the event-split regime, the models are also trained on different watermarking keys at the same time, simulating a realistic aspect of the deployment scenario where different sub-datasets may be protected with different watermark keys within a single training corpus.

\section{Extended Experimental Setup Details}\label{app:ext-exp-details}

\paragraph{Watermark Scheme Details}\label{app:wm-detail}

All four experimental families (event-split finetuning, SKS finetuning, CPT pretraining, and from-scratch pretraining) share the same KGW-style watermark configuration: greenlist fraction $\gamma = 0.5$, logit bias $\delta = 4.0$, and a 2-token prefix used to seed the PRF that selects each step's greenlist. Headline significance values come from the empirical-null permutation test of \cref{sec:p-value-validity}.

Detection is run in teacher-forced \emph{reading} mode: the target model is run over the tokenized source text and scored at each position under the preceding tokens of that same source text, not on free generations. We score under argmax mode---counting positions where the model's greedy next-token lands inside the keyed greenlist. Repeated prefix windows are deduplicated by context so that high-frequency local contexts cannot dominate either the watermark score or the empirical null. For the empirical null, we draw $M = 10^6$ randomized keys per surface from a fixed integer keyspace; this gives an exact $p$-value resolution of $10^{-6}$, with an empirical-Gaussian fit to the same null distribution available as an extrapolated reference where smaller $p$-values are needed.

The watermark keys themselves are scaffolded differently across the three experimental settings. The event-split finetuning grid uses, at each $F$ level, a fresh set of $F$ fold-specific keys (one per fold), and the detector key for a given watermark surface is always matched to the fold ID being read. The SKS ablation does not use fold-specific keys; it uses four fixed keys, each held constant across the entire $(F, E)$ grid for one of four model families, with the detector key matched to the family's single key. Phase 2 pretraining (both CPT and from-scratch) reuses the same two $F=2$ fold keys from the Phase 1 event-split anchor row across all ten schedules and both initialization regimes; this is deliberate cross-phase calibration, which is also why the warm-key handedness present in the Phase 1 $F=2$ scaffold (\cref{sec:appendix-phase1-event-null-validity}) is visible in the Phase 2 keyed readouts.

\paragraph{Data Folding.}\label{app:data-fold-exp-detail}

For each finetuning experiment, we define a fold factor $F$ and partition the FictionalQA documents into $F$ equal folds. We then train a small population of models per cell with paired clean and watermarked twins, so that each watermarked positive has a matched clean ``false-probe'' negative trained on the unmarked version of the same documents. To simulate the watermark-based dataset protection approach, for every fold $D_i$, we create a paraphrased version of the documents using a paraphraser model running the watermark decoding scheme with key $k_i$, producing $D_{k_i} = T_m(D_i, k_i)$. The watermarked model $f^{wm}_{\theta_i}$ is trained on a mixture that includes $D_{k_i}$ and other data, while the corresponding clean twin $f^{c}_{\theta_i}$ trains on $D_i$ in its original form before the paraphrasing transformation. We also vary the number of effective epochs $E$ over the watermarked subset within a run while holding the per-cell training-token budget fixed across $(F, E)$; samples from the base data source are not repeated. \Cref{tab:phase1-sks-idealized-exposure} summarizes the resulting effective per-key exposure across the grid; the event-split and SKS regimes share this idealized table.

\section{Finetuning Event-Split Exhaustive Readout}\label{sec:appendix-phase1-event}

This appendix carries the full per-cell readout of the event-split finetuning grid that the main body's \cref{fig:event-keyed-aligned-exact} introduces. The main body uses the empirical-exact reference for both the keyed-signal and DIA heatmaps; the subsections below add the empirical-Gaussian companion of each, the packed-surface counterparts of the keyed/null pair, the cross-key sham-null heatmaps, the realized exposure and $\hat{E}$ tables, the row-level MIA table for the baselines, and the null-validity panel.

\subsection{Empirical-Gaussian and Packed-Surface Heatmap Companions}\label{sec:appendix-phase1-event-heatmaps}

\begin{figure}[h!]
    \centering
    \includegraphics[width=0.49\textwidth]{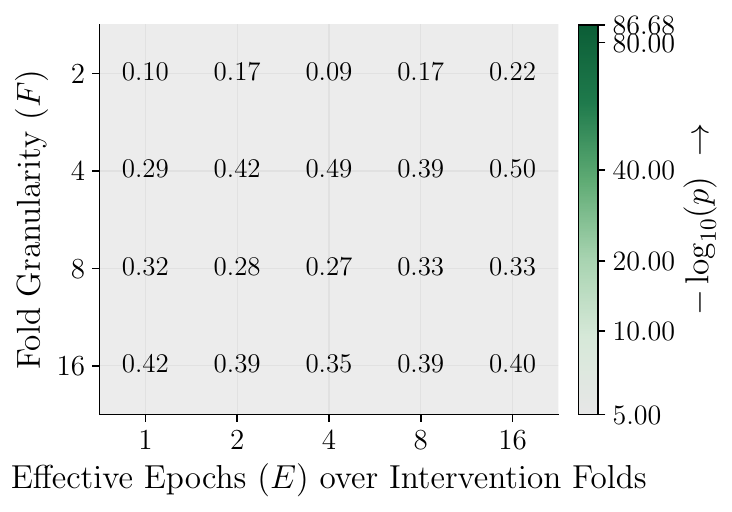}
    \hfill
    \includegraphics[width=0.49\textwidth]{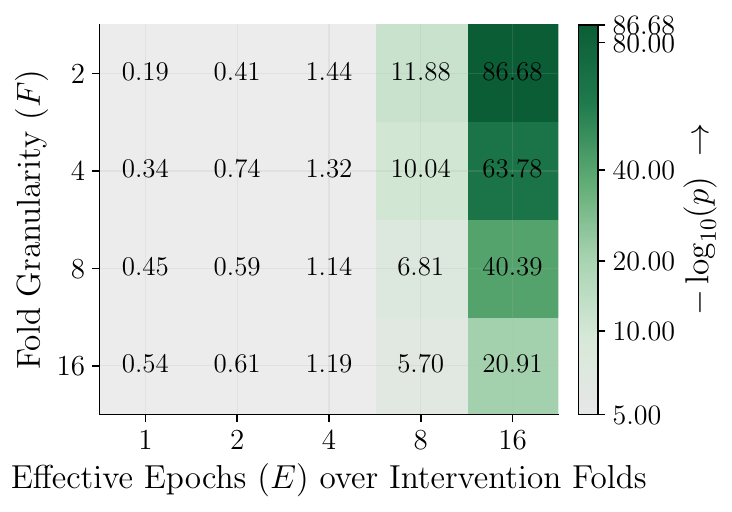}
    \caption{Finetuning event-split matched clean-twin false-probe null (left) and keyed signal (right) across the $F \times E$ grid on the aligned unpacked detection surface, scored as $-\log_{10} p$ under the empirical-Gaussian reference.}
    \label{fig:appendix-event-keyed-aligned-gaussian}
\end{figure}

\begin{figure}[h!]
    \centering
    \includegraphics[width=0.49\textwidth]{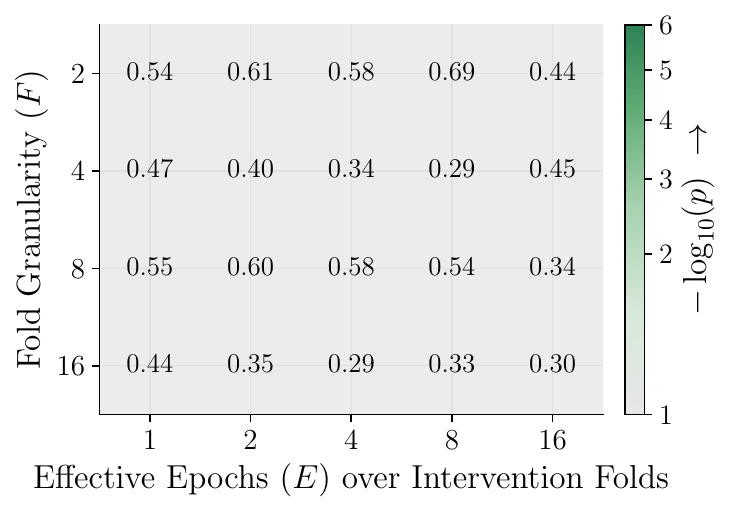}
    \hfill
    \includegraphics[width=0.49\textwidth]{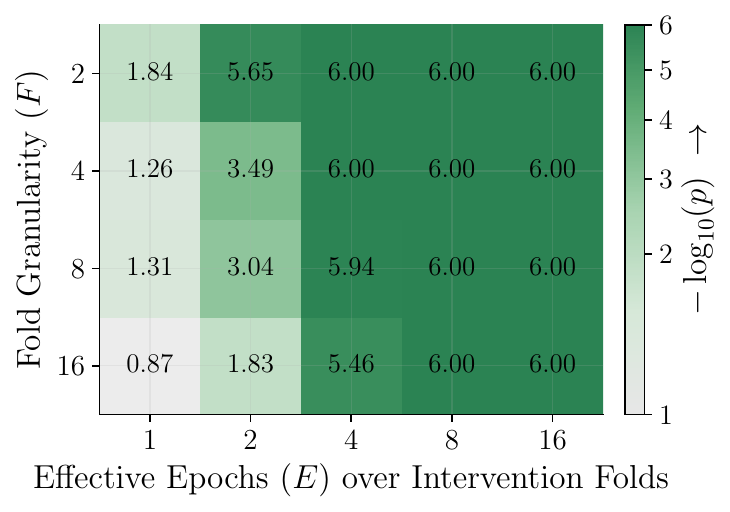}
    \caption{Finetuning event-split matched clean-twin false-probe null (left) and keyed signal (right) across the $F \times E$ grid on the packed detection surface, scored as $-\log_{10} p$ under the empirical-exact null. The packed surface is a more permissive oracle baseline than the aligned surface.}
    \label{fig:appendix-event-keyed-packed-exact}
\end{figure}

\begin{figure}[h!]
    \centering
    \includegraphics[width=0.49\textwidth]{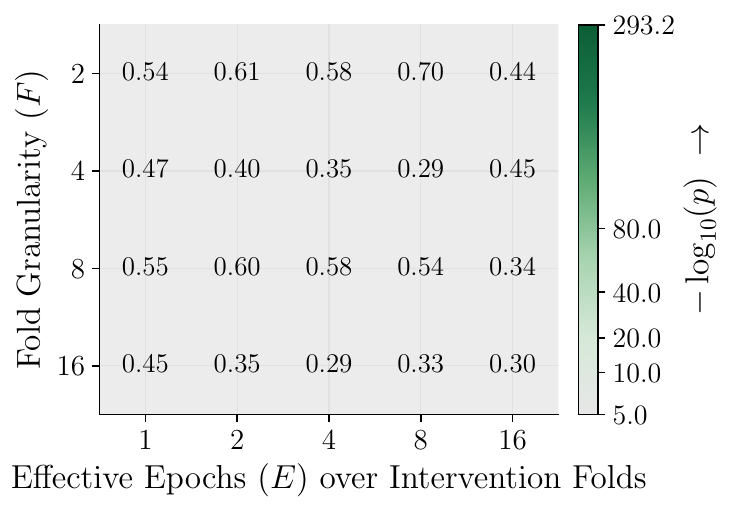}
    \hfill
    \includegraphics[width=0.49\textwidth]{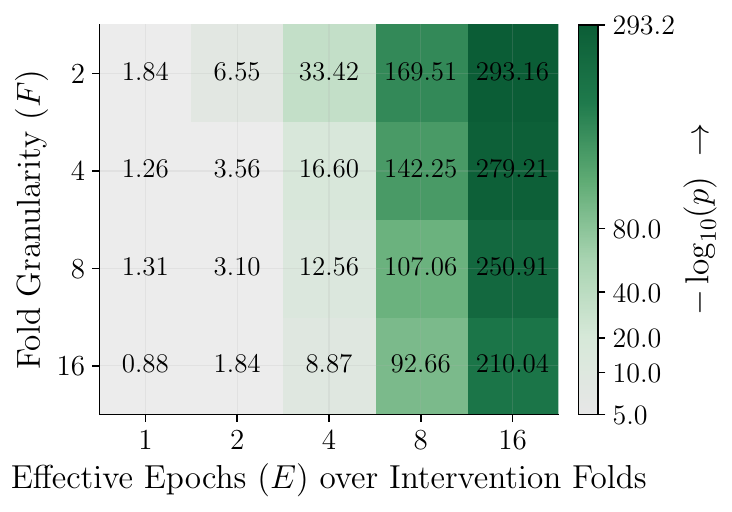}
    \caption{Finetuning event-split matched clean-twin false-probe null (left) and keyed signal (right) across the $F \times E$ grid on the packed detection surface, scored as $-\log_{10} p$ under the empirical-Gaussian reference.}
    \label{fig:appendix-event-keyed-packed-gaussian}
\end{figure}

\begin{figure}[h!]
    \centering
    \includegraphics[width=0.49\textwidth]{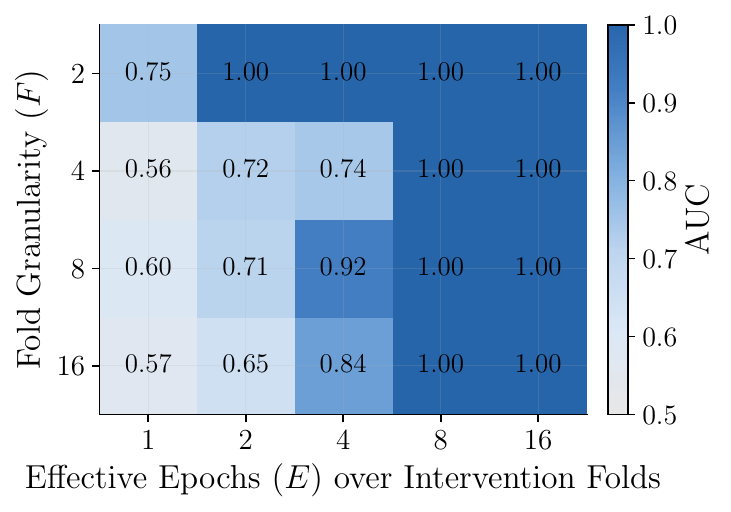}
    \hfill
    \includegraphics[width=0.49\textwidth]{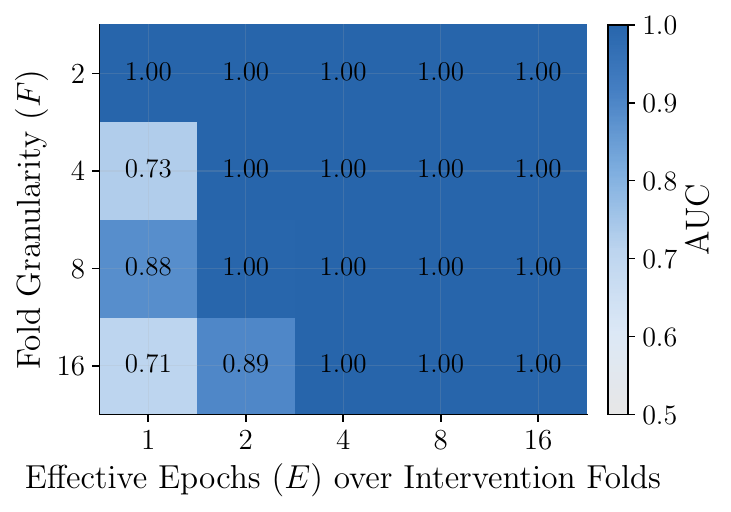}
    \caption{Finetuning event-split watermark whole-model DIA AUC across the $F \times E$ grid, on the aligned (left) and packed (right) detection surfaces, both scored against an empirical-Gaussian null.}
    \label{fig:appendix-event-dia-gaussian}
\end{figure}

\Cref{fig:event-keyed-aligned-exact,fig:appendix-event-keyed-aligned-gaussian,fig:appendix-event-keyed-packed-exact,fig:appendix-event-keyed-packed-gaussian} carry the four event-split keyed/null pairs across the (aligned, packed) surface and (exact, Gaussian) $p$-value-type combinations; the aligned-exact keyed half is also shown in the main-body \cref{fig:event-keyed-aligned-exact}. \Cref{fig:event-dia-exact,fig:appendix-event-dia-gaussian} carry the matching watermark whole-model DIA AUC pairs, with aligned and packed surfaces shown side-by-side under each $p$-value type.

\subsection{Event-Split Exposure Trend Curves}\label{sec:appendix-phase1-event-trend}

\begin{figure}[h!]
    \centering
    \includegraphics[width=0.95\textwidth]{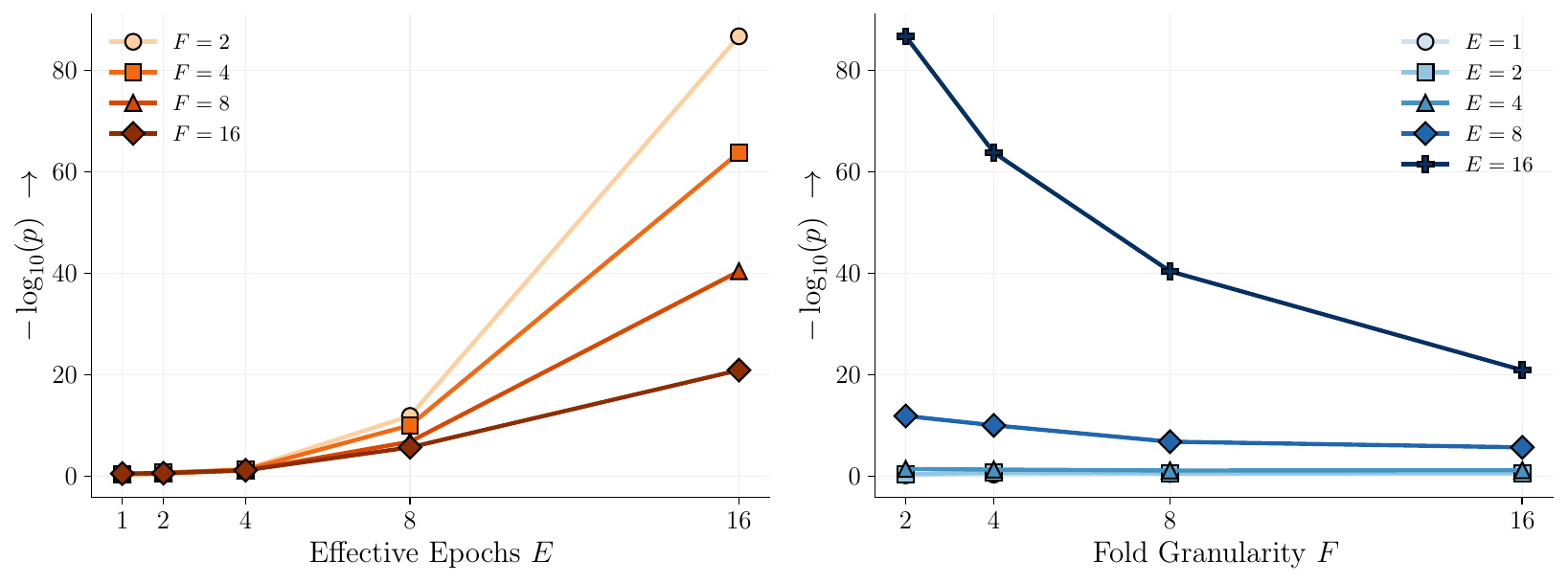}
    \caption{Finetuning event-split keyed-signal exposure response across the grid, tracing $-\log_{10} p$ as a function of effective per-key exposure with separate lines for each fold count $F$. This figure plots the same per-cell keyed-signal values as the right panel of \cref{fig:event-keyed-aligned-exact}, just computed with the empirical Gaussian extrapolated p-values from \cref{fig:appendix-event-keyed-aligned-gaussian} and re-cast on a continuous exposure axis to make the trend clearer.}
    \label{fig:event-trend}
\end{figure}

\Cref{fig:event-trend} is a re-visualization of the same keyed-signal data shown in the right panel of \cref{fig:event-keyed-aligned-exact}, but using the empirical Gaussian extrapolation p-values from \cref{fig:appendix-event-keyed-aligned-gaussian} and re-cast on a continuous exposure axis with separate lines per $F$ to make the exposure trend and the cross-$F$ separation easier to read at a glance. At low exposure the per-key support is too small for keyed signal to lift off in any of the $F$ rows, while at high exposure the curves separate cleanly with $F$.

\subsection{Cross-Key Sham-Null Heatmaps}\label{sec:appendix-phase1-event-sham-null}

\begin{figure}[h!]
    \centering
    \includegraphics[width=0.49\textwidth]{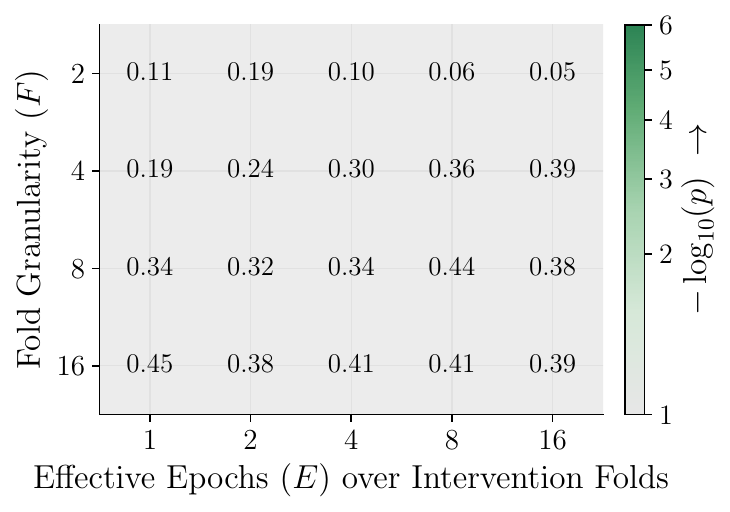}
    \hfill
    \includegraphics[width=0.49\textwidth]{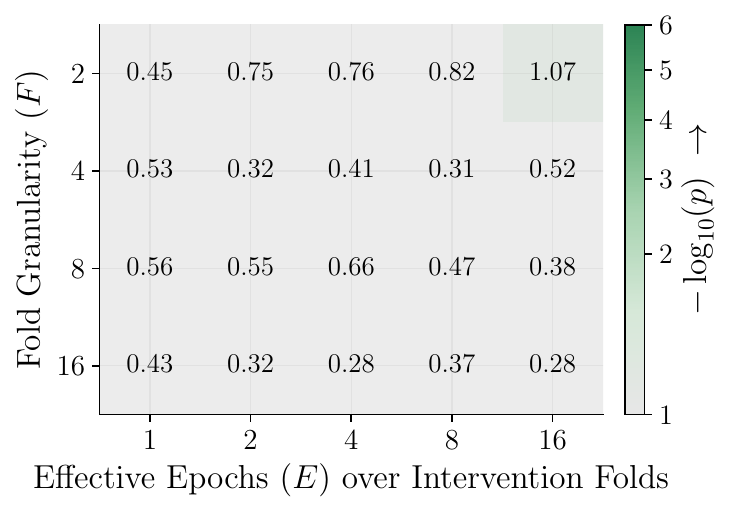}
    \caption{Finetuning event-split cross-key sham-null heatmap across the $F \times E$ grid, on the aligned (left) and packed (right) detection surfaces, scored as $-\log_{10} p$ under the empirical-exact null. The watermark detector is queried with a key the target model never saw in training, providing a negative control beyond the matched clean-twin false-probe null.}
    \label{fig:appendix-event-sham-null-exact}
\end{figure}

\begin{figure}[h!]
    \centering
    \includegraphics[width=0.49\textwidth]{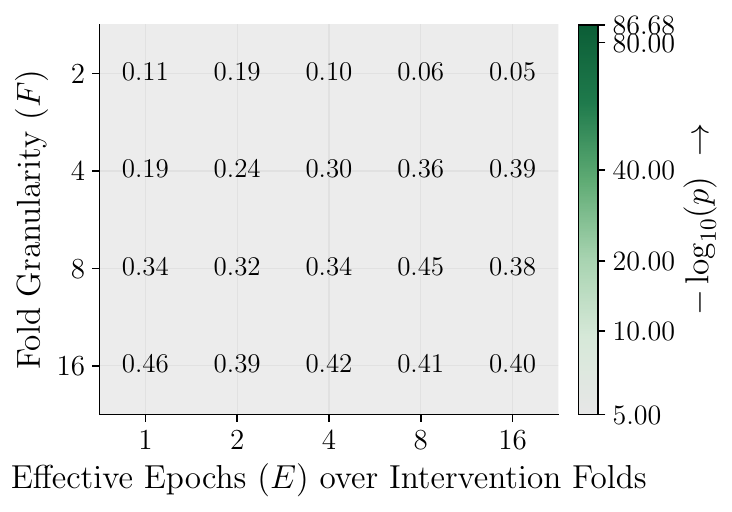}
    \hfill
    \includegraphics[width=0.49\textwidth]{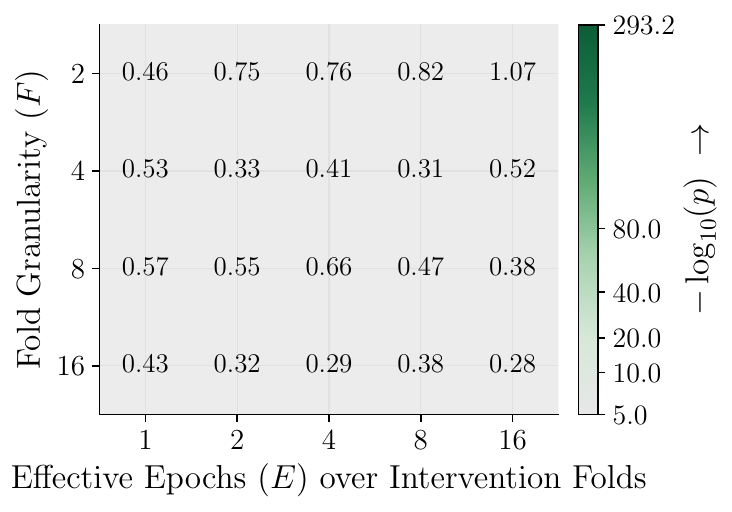}
    \caption{Finetuning event-split cross-key sham-null heatmap across the $F \times E$ grid, on the aligned (left) and packed (right) detection surfaces, scored as $-\log_{10} p$ under the empirical-Gaussian null.}
    \label{fig:appendix-event-sham-null-gaussian}
\end{figure}

\Cref{fig:appendix-event-sham-null-exact,fig:appendix-event-sham-null-gaussian} carry the cross-key sham-null heatmaps for the event-split grid, where the watermark detector is queried with a key the target model never saw in training. These act as an additional negative control beyond the matched clean-twin false-probe null, isolating per-key idiosyncrasy of the detection surface from the matched-pair design.

\subsection{Event-Split Realized Exposure and $\hat{E}$ Per-Cell Readbacks}\label{sec:appendix-phase1-event-ehat}

\Cref{tab:wm4memb-phase1-event-grid-fixed-exposure-current} reports the realized normalized exposure $\hat{E}/F$ per cell of the event-split finetuning grid. \Cref{tab:wm4memb-phase1-event-grid-fixed-ehat-current} reports the underlying realized $\hat{E}$ values. The corresponding idealized epoch counts $E$ are the column headers of \cref{tab:phase1-sks-idealized-exposure}, which the event-split construction shares with SKS.

\input{auto_figures/wm4memb_phase1_event_grid_fixed_exposure_current.tex}

\input{auto_figures/wm4memb_phase1_event_grid_fixed_ehat_current.tex}

\subsection{Event-Split Training Scale and Trial Geometry}\label{sec:appendix-phase1-event-context}

\Cref{tab:wm4memb-phase1-event-fixed-training-scale-context} reports the per-cell watermarked-token totals (mean, min--max across paired models) and the corresponding fraction of each model's 131M-token training budget. \Cref{tab:wm4memb-phase1-event-fixed-dia-context} reports the per-cell paired-model counts and the resulting $n_+ / n_-$ trial counts that drive each cell's whole-model DIA AUC; both are summarized in \cref{sec:phase1}.

\input{auto_figures/wm4memb_phase1_event_grid_fixed_training_scale_context_current.tex}

\input{auto_figures/wm4memb_phase1_event_grid_fixed_dia_context_current.tex}

\subsection{Event-Split Loss-Based and Reference-Model Baselines Row-Level MIA}\label{sec:appendix-phase1-event-baselines}

\input{auto_figures/wm4memb_phase1_event_grid_fixed_row_baselines_current.tex}

\begin{table*}[h!]
\centering
\caption{To contextualize the performance differences between the watermark-based DIA technique and other baselines presented in \cref{tab:wm4memb-phase1-event-grid-fixed-dia-current}, here we summarize the key assumptions about the data owner's (i.e. attacker's) control over the data and access to the target model needed to make a dataset membership claim. \textbf{Qualifications:} $^1$The ``reading-mode'' method of \citet{sander2024watermarking,sander2025detecting} requires the ability to check what the next-token prediction of the target model would be conditioned on a prefix of the protected sample. While in theory this could be accomplished by repeatedly querying the model while iteratively appending tokens to the prompt (expensive), the practicality of this operation given modern frontier model APIs is questionable.$^2$We mark this row as such to draw attention to the fact that performing academic analyses centering ROC-AUC allows us to ignore the necessity of calibrating a single decision threshold to deploy. However, in practice, for all MIA/DIA techniques that do not admit a p-value or other analytic (FP) error estimate, held-out validation samples must be used to calibrate the decision threshold.}
\label{tab:access-assumptions}
\begin{tabular}{m{4cm}ccccccc}
\hline
Access Assumption & Watermarked & Raw-loss & Argmax & min-k$_{10}$ & rMIA-simple & rMIA & LiRA \\
\midrule
Data owner can modify their data before it is included in the training dataset. & \checkemoji{} &  &  &  &  &  & \\\hline
Data owner has next-token prediction$^1$ access to target model.  & \checkemoji{} &  & \checkemoji{} &  &  &  &  \\\hline
Data owner must maintain held-out pool of additional samples for calibration$^2$.  & & \checkemoji{} & \checkemoji{} & \checkemoji{} & \checkemoji{} & \checkemoji{} & \checkemoji{} \\\hline
Data owner has loss or logit access to target model.  &  & \checkemoji{} &  & \checkemoji{} & \checkemoji{} & \checkemoji{} & \checkemoji{} \\\hline
Data owner has compute and target model details needed to train reference models.  &  &  &  &  & \checkemoji{} & \checkemoji{} & \checkemoji{} \\\hline
\end{tabular}
\end{table*}

\subsection{Event-Split Null-Validity Extended}\label{sec:appendix-phase1-event-null-validity}

\begin{figure}[ht!]
    \centering
    \includegraphics[width=0.95\textwidth]{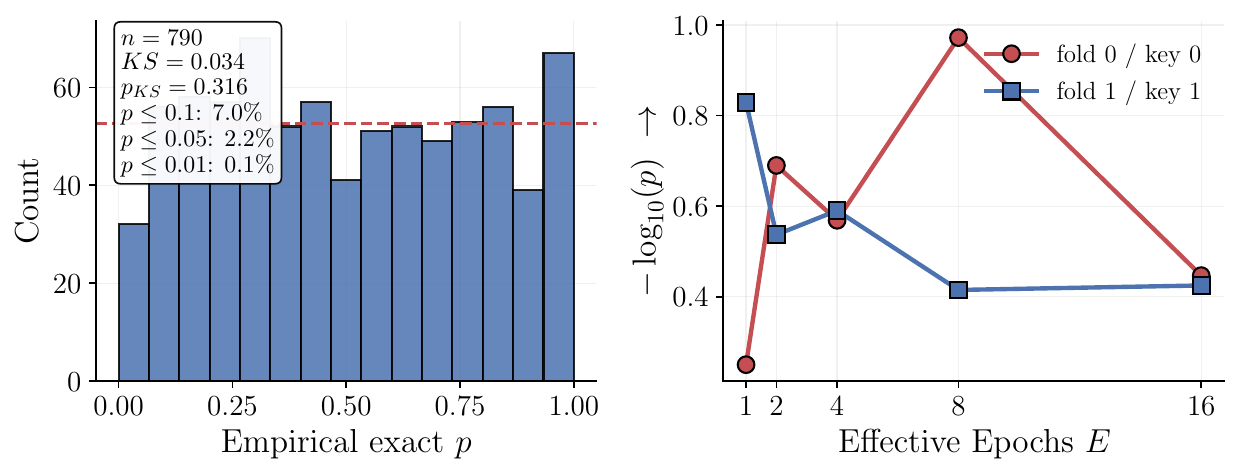}
    \caption{Event-grid null-validity panel. \textbf{Left:} histogram of pooled empirical exact $p$-values from the packed matched-clean-negative whole-model readings across the $(F,E)$ event grid. The annotated statistic and $p$-value are from a one-sample Kolmogorov-Smirnov test of these pooled $p$-values against $\mathrm{Uniform}(0, 1)$; the dashed horizontal line is the expected per-bin count under a uniform histogram with the plotted binning. \textbf{Right:} $-\log_{10} p$ trace of the $F=2$ warm-key slice of the same packed null family plotted against $E$, shown for context only and not part of the KS test on the left.}
    \label{fig:appendix-event-null-validity}
\end{figure}

The clipped chart \Cref{fig:event-null-validity} in the main body, equivalent to the left side of \cref{fig:appendix-event-null-validity} confirms that once pooled across many distinct positive keys, the empirical exact null is close to uniform at the 1M-null scale and respects standard tail-rate thresholds, supporting the use of the empirical-null permutation test as the headline reference distribution for the keyed readout. Per-key idiosyncrasy is a separate concern visible in the right-hand trace: at very small fold counts a single key can run systematically warm even when the pooled null is well-calibrated, and the inherited $F=2$ scaffold reused in pretraining (\cref{sec:phase2}) carries one such warm key.

\section{Finetuning SKS Exhaustive Readout}\label{sec:appendix-phase1-sks}

This appendix reports the simple per-key support (SKS) ablation, where each model trains on exactly one watermarked fold so the per-model fictional support fraction shrinks as $1/F$ instead of being held fixed across $F$. SKS is structurally simpler than the event-split regime (\cref{sec:appendix-phase1-event}) but unrealistic in that no real deployed corpus would protect only one fold of one data owner's content under a single key. The same idealized per-key exposure (\cref{tab:phase1-sks-idealized-exposure}) governs both regimes; what differs is sibling support and the per-cell positive/negative trial budget (\cref{tab:wm4memb-phase1-sks-dia-context}). We use SKS as a watermark-only sanity check on the per-key scaling story without sibling-key interference, and run the loss-based and reference-model comparison only on the more realistic event-split regime where the row-level baselines are meaningful.

\subsection{SKS Empirical-Gaussian and Packed-Surface Heatmap Companions}\label{sec:appendix-phase1-sks-heatmaps}

\begin{figure}[h!]
    \centering
    \includegraphics[width=0.49\textwidth]{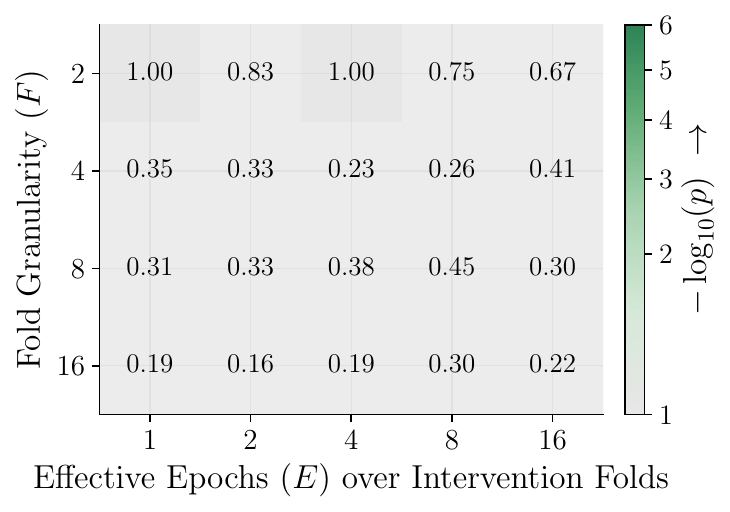}
    \hfill
    \includegraphics[width=0.49\textwidth]{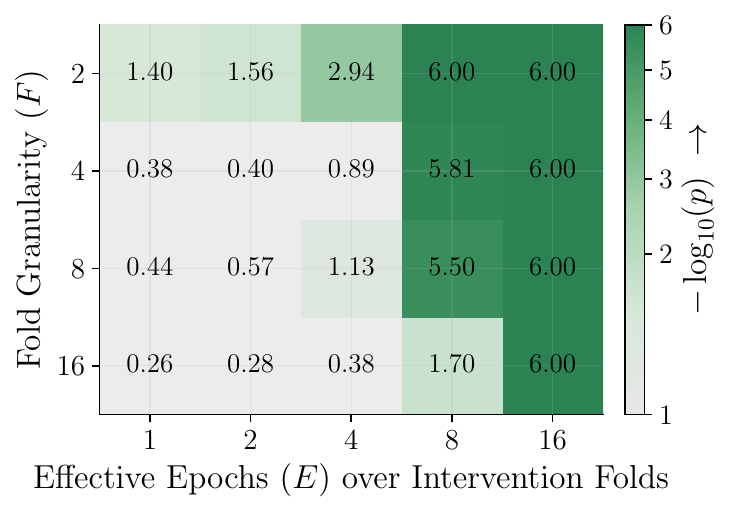}
    \caption{Finetuning SKS matched clean-twin false-probe null (left) and keyed signal (right) across the $F \times E$ grid on the aligned unpacked detection surface, scored as $-\log_{10} p$ under an empirical-exact reference. The false-probe null stays quiet on the same surface, validating the matched clean-twin negative as the right baseline against which to read the keyed map; the keyed signal grows monotonically with $E$ and decays with $F$ as the per-key support fraction $1/F$ shrinks.}
    \label{fig:sks-heatmaps}
\end{figure}

\begin{figure}[h!]
    \centering
    \includegraphics[width=0.49\textwidth]{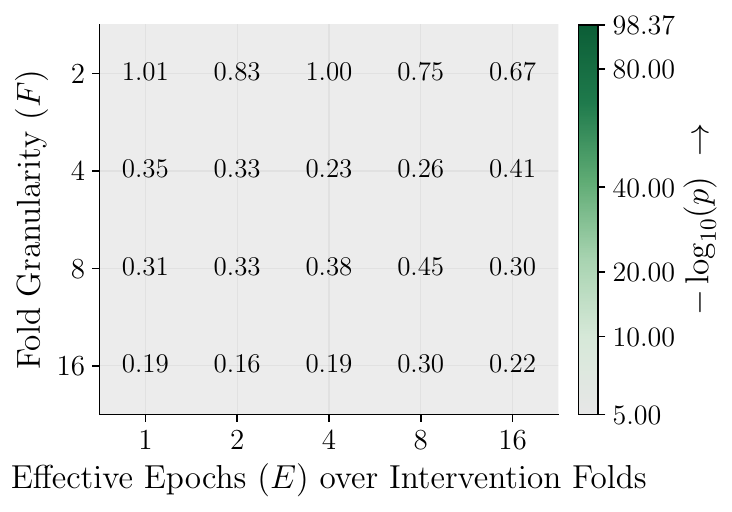}
    \hfill
    \includegraphics[width=0.49\textwidth]{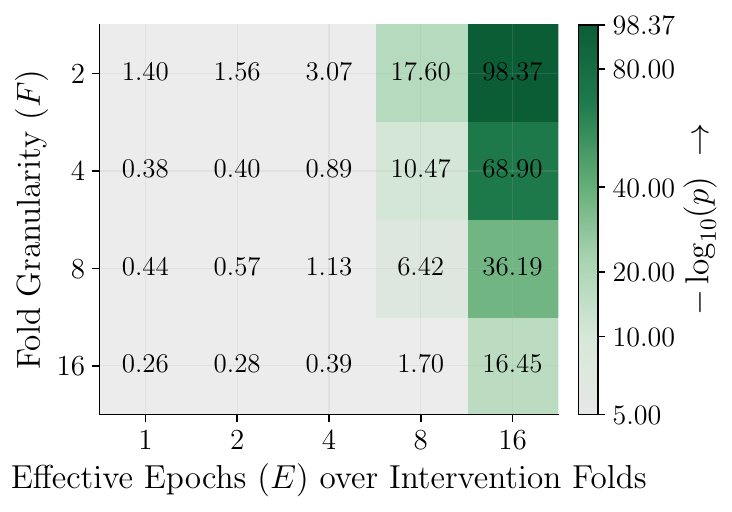}
    \caption{Finetuning SKS matched clean-twin false-probe null (left) and keyed signal (right) across the $F \times E$ grid on the aligned unpacked detection surface, scored as $-\log_{10} p$ under the empirical-Gaussian reference.}
    \label{fig:appendix-sks-keyed-gaussian-aligned}
\end{figure}

\begin{figure}[h!]
    \centering
    \includegraphics[width=0.49\textwidth]{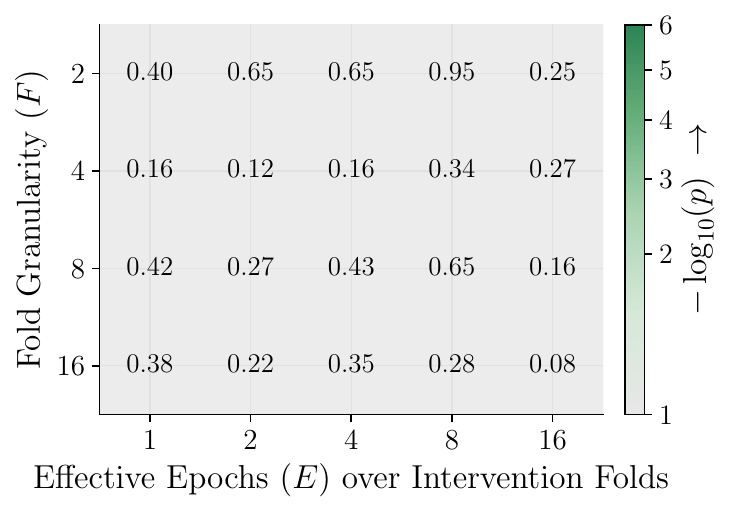}
    \hfill
    \includegraphics[width=0.49\textwidth]{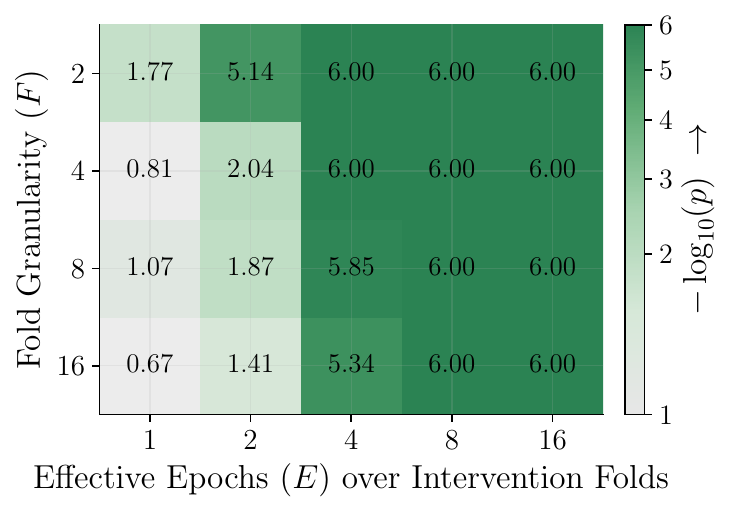}
    \caption{Finetuning SKS matched clean-twin false-probe null (left) and keyed signal (right) across the $F \times E$ grid on the packed detection surface, scored as $-\log_{10} p$ under the empirical-exact null. The packed surface is a more permissive oracle baseline than the aligned surface.}
    \label{fig:appendix-sks-keyed-exact-packed}
\end{figure}

\begin{figure}[h!]
    \centering
    \includegraphics[width=0.49\textwidth]{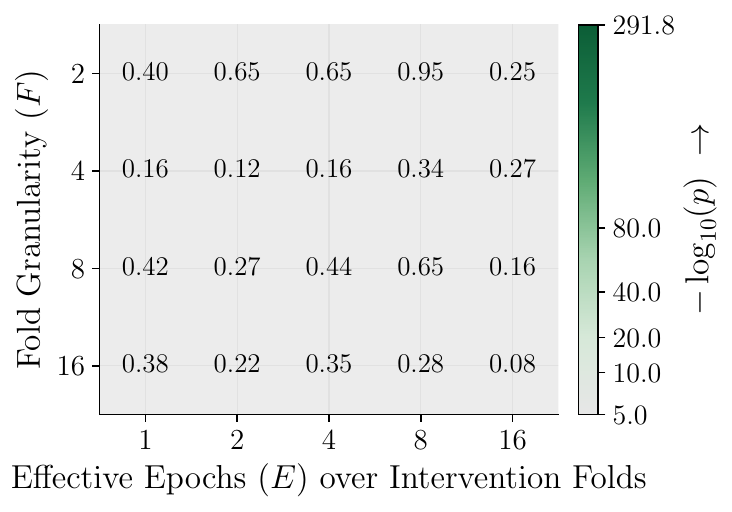}
    \hfill
    \includegraphics[width=0.49\textwidth]{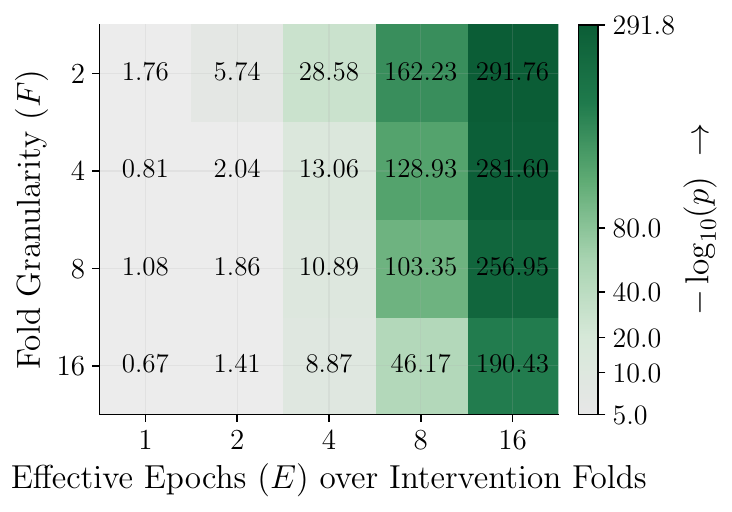}
    \caption{Finetuning SKS matched clean-twin false-probe null (left) and keyed signal (right) across the $F \times E$ grid on the packed detection surface, scored as $-\log_{10} p$ under the empirical-Gaussian reference.}
    \label{fig:appendix-sks-keyed-gaussian-packed}
\end{figure}

\begin{figure}[h!]
    \centering
    \includegraphics[width=0.49\textwidth]{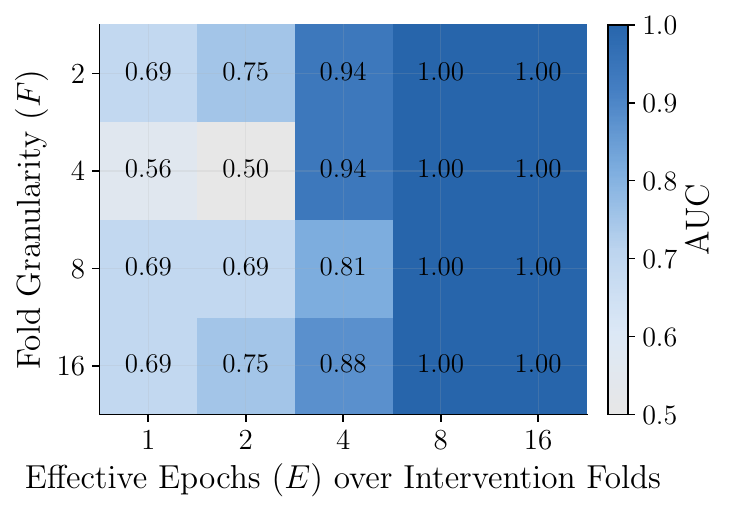}
    \hfill
    \includegraphics[width=0.49\textwidth]{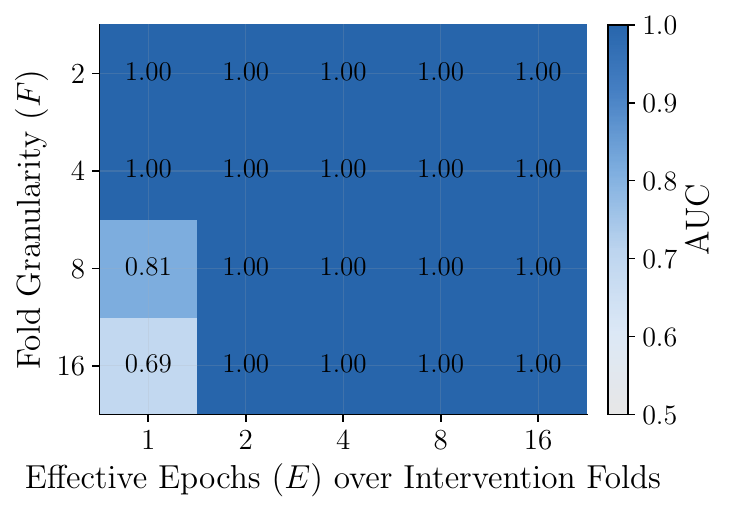}
    \caption{Finetuning SKS watermark whole-model DIA AUC across the $F \times E$ grid, on the aligned (left) and packed (right) detection surfaces, both scored against an empirical-exact null. On the aligned surface the AUC saturates at $1.0$ from $E=8$ onward at every $F$, while the lowest-$E$ corner remains coarse with only eight trials per cell. On the packed surface, the more permissive oracle recovers several of those low-exposure cells, reaching $1.0$ one to two $E$-steps earlier across the grid.}
    \label{fig:sks-dia-heatmaps}
\end{figure}

\begin{figure}[h!]
    \centering
    \includegraphics[width=0.49\textwidth]{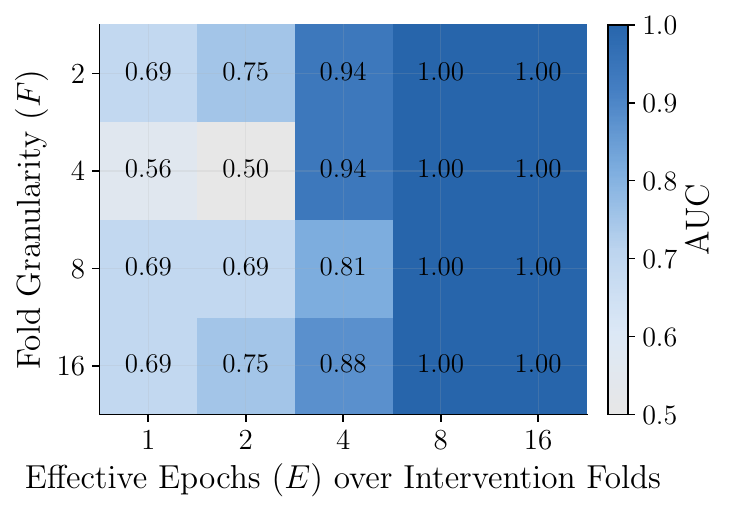}
    \hfill
    \includegraphics[width=0.49\textwidth]{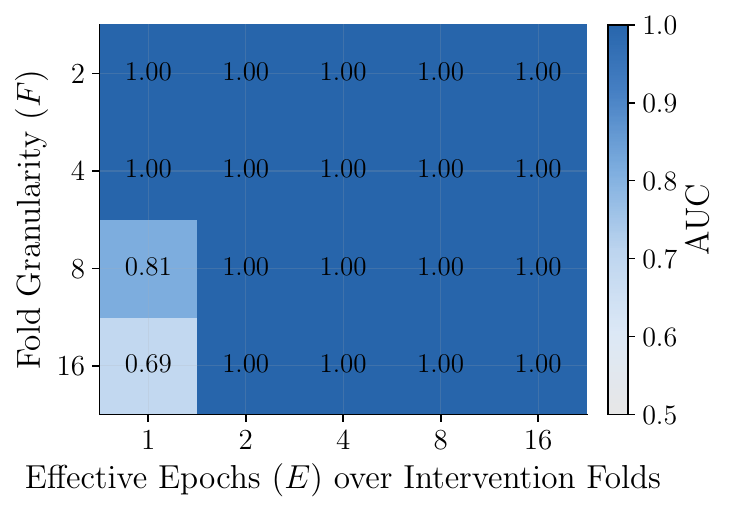}
    \caption{Finetuning SKS watermark whole-model DIA AUC across the $F \times E$ grid, on the aligned (left) and packed (right) detection surfaces, both scored against an empirical-Gaussian null.}
    \label{fig:appendix-sks-dia-gaussian}
\end{figure}

\Cref{fig:sks-heatmaps,fig:sks-dia-heatmaps,fig:appendix-sks-keyed-gaussian-aligned,fig:appendix-sks-keyed-gaussian-packed,fig:appendix-sks-keyed-exact-packed,fig:appendix-sks-dia-gaussian} carry the SKS keyed/null pair and DIA AUC heatmap pair on the aligned-exact, aligned-Gaussian, packed-exact, and packed-Gaussian detection surfaces.

\subsection{SKS Exposure Trend Curves}\label{sec:appendix-phase1-sks-trend}

\begin{figure}[h!]
    \centering
    \includegraphics[width=0.95\textwidth]{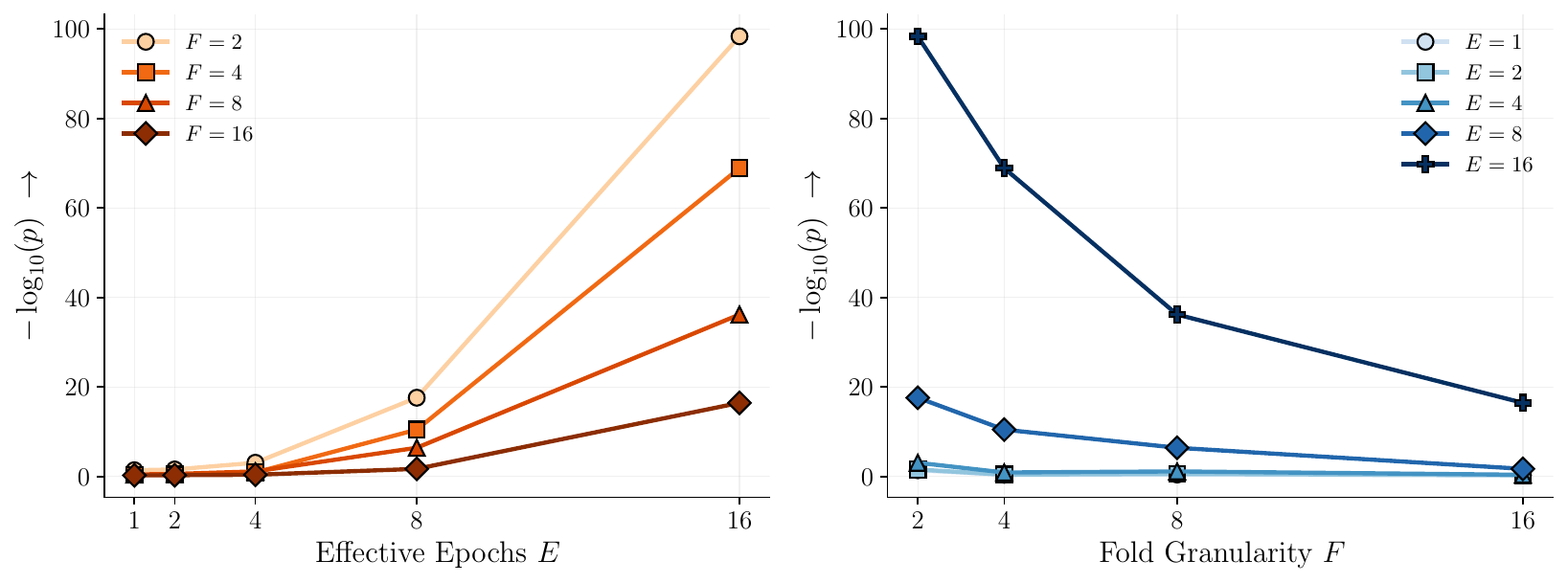}
    \caption{Finetuning SKS keyed-signal exposure response across the grid, tracing $-\log_{10} p$ as a function of effective per-key exposure with separate lines for each fold count $F$. This figure plots the same per-cell keyed-signal values as the right panel of \cref{fig:sks-heatmaps}, just re-cast on a continuous exposure axis to make the trend clearer.}
    \label{fig:sks-trend}
\end{figure}

\Cref{fig:sks-trend} is a re-visualization of the same keyed-signal data shown in the right panel of \cref{fig:sks-heatmaps}, re-cast on a continuous exposure axis with separate lines per $F$ to make the exposure trend and the cross-$F$ separation easier to read at a glance: at low exposure the per-key support is too small for keyed signal to lift off in any of the $F$ rows, while at high exposure the curves separate cleanly with $F$.

\subsection{SKS Realized Exposure and $\hat{E}$ Per-Cell Readbacks}\label{sec:appendix-phase1-sks-ehat}

\Cref{tab:wm4memb-phase1-sks-exposure-current} reports the realized normalized exposure $\hat{E}/F$ per cell of the finetuning SKS grid, the direct realized counterpart to the idealized $E/F$ values in \cref{tab:phase1-sks-idealized-exposure}. \Cref{tab:wm4memb-phase1-sks-ehat-current} reports the underlying realized $\hat{E}$ values. The discrepancy between idealized and realized values here is the realized overshoot of the watermarked subset's effective epoch count relative to the planned schedule.

\input{auto_figures/wm4memb_phase1_sks_exposure_current.tex}

\input{auto_figures/wm4memb_phase1_sks_ehat_current.tex}

\subsection{SKS Training Scale and Trial Geometry}\label{sec:appendix-phase1-sks-context}

\Cref{tab:wm4memb-phase1-sks-training-scale-context} reports the per-cell watermarked-token totals (mean, min--max across paired models) and the corresponding fraction of each model's 131M-token training budget. \Cref{tab:wm4memb-phase1-sks-dia-context} reports the per-cell paired-model counts and the resulting $n_+ / n_-$ trial counts; SKS holds the per-cell trial budget fixed at $4 / 4$ across the grid, since each model trains on exactly one watermarked fold and the per-cell positive/negative budget is the same at every $(F, E)$.

\input{auto_figures/wm4memb_phase1_sks_training_scale_context_current.tex}

\input{auto_figures/wm4memb_phase1_sks_dia_context_current.tex}

\subsection{SKS Watermark-Only DIA Numeric Table}\label{sec:appendix-phase1-sks-baselines}

\Cref{tab:wm4memb-phase1-sks-watermark-dia-current} carries the underlying numeric AUC values that drive the SKS DIA heatmap pair in \cref{fig:sks-dia-heatmaps}. We deliberately keep this table watermark-only: the SKS support construction is a clean per-key scaling ablation in which each model trains on exactly one watermarked fold, and the loss-based / reference-model row-level baselines are only meaningful against the realistic event-split regime where multiple keys coexist, so the head-to-head comparison against those baselines is run only there (\cref{sec:appendix-phase1-event-baselines}).

\input{auto_figures/wm4memb_phase1_sks_watermark_dia_current.tex}

\subsection{SKS Null-Validity}\label{sec:appendix-phase1-sks-null-validity}

\begin{figure}[h!]
    \centering
    \includegraphics[width=0.95\textwidth]{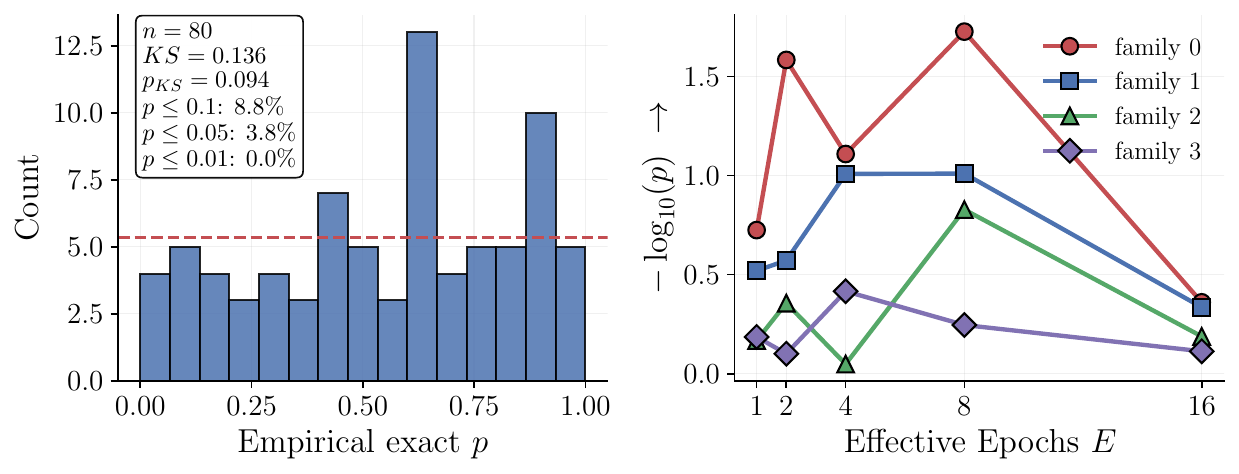}
    \caption{SKS null-validity panel. \textbf{Left:} histogram of pooled empirical exact $p$-values from the packed clean-model watermark-surface false-probe rows across the depth-4 SKS grid. The annotated statistic and $p$-value are from a one-sample Kolmogorov-Smirnov test of these pooled $p$-values against $\mathrm{Uniform}(0, 1)$; the dashed horizontal line is the expected per-bin count under a uniform histogram with the plotted binning. \textbf{Right:} $-\log_{10} p$ trace of the $F=2$ warm-key slice of the same packed null family plotted against $E$, shown for context only and not part of the KS test on the left.}
    \label{fig:sks-null-validity}
\end{figure}

\Cref{fig:sks-null-validity} confirms that, once pooled across many distinct positive keys, the empirical exact null is close to uniform at the 1M-null scale and respects standard tail-rate thresholds. This holds in both the event-split grid (\cref{fig:event-null-validity}) and in this SKS ablation, supporting the use of the empirical-null permutation test as the headline reference distribution across both regimes.

\section{Pretraining Exhaustive Readout}\label{sec:appendix-phase2-dia}

This appendix carries the full per-init readout of the pretraining schedule sweep to complement the main body's \cref{fig:phase2-cpt-aligned-exact,fig:phase2-scratch-aligned-exact}. The structure is grouped first by initialization regime (CPT then from-scratch), and within each by aligned-then-packed surface, exact-then-Gaussian $p$-value type, with the watermark whole-model DIA bar pairs trailing each init's keyed/null block. The split row-level MIA and whole-model DIA baseline tables for both initialization regimes are reported below the per-init bar companions, and each whole-model DIA cell is computed over $2{+}/2{-}$ trials per schedule.

\begin{table}[h!]
\centering
\caption{Pretraining idealized exposure profile across the ten-schedule sweep at $F=2$. Each schedule targets a nominal effective epoch count $E$, which at $F=2$ corresponds to an idealized per-key exposure $E/F = E/2$. Both initialization regimes (CPT and from-scratch) share the same idealized profile.}
\label{tab:phase2-idealized-exposure}
\begin{tabular}{lcc}
\hline
Schedule group & $E$ & $E/F$ \\
\hline
(S1,E1) -- (S4,E1) & 1 & 0.5 \\
(U,E4) / (P,E4) & 4 & 2.0 \\
(U,E8) / (P,E8) & 8 & 4.0 \\
(U,E16) / (P,E16) & 16 & 8.0 \\
\hline
\end{tabular}
\end{table}

\subsection{Pretraining CPT Companion Figures}\label{sec:appendix-phase2-cpt-companions}

\begin{figure}[h!]
    \centering
    \includegraphics[width=0.49\textwidth]{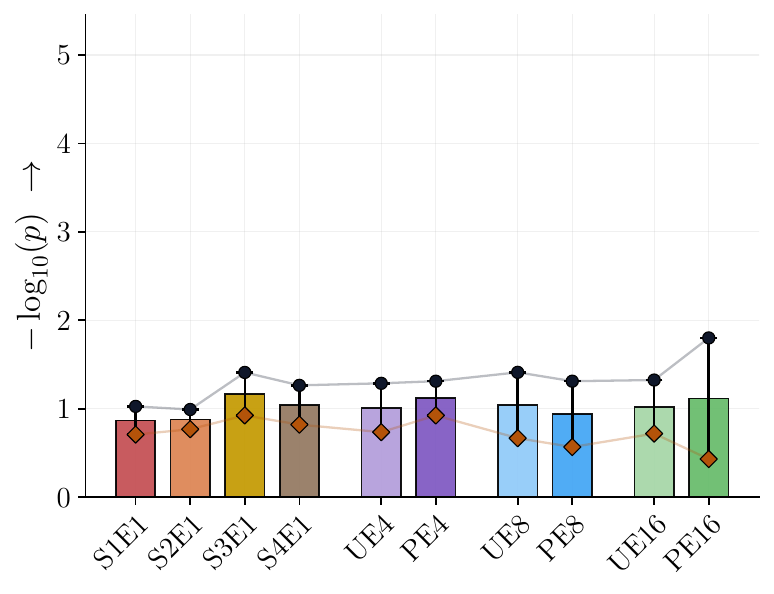}
    \hfill
    \includegraphics[width=0.49\textwidth]{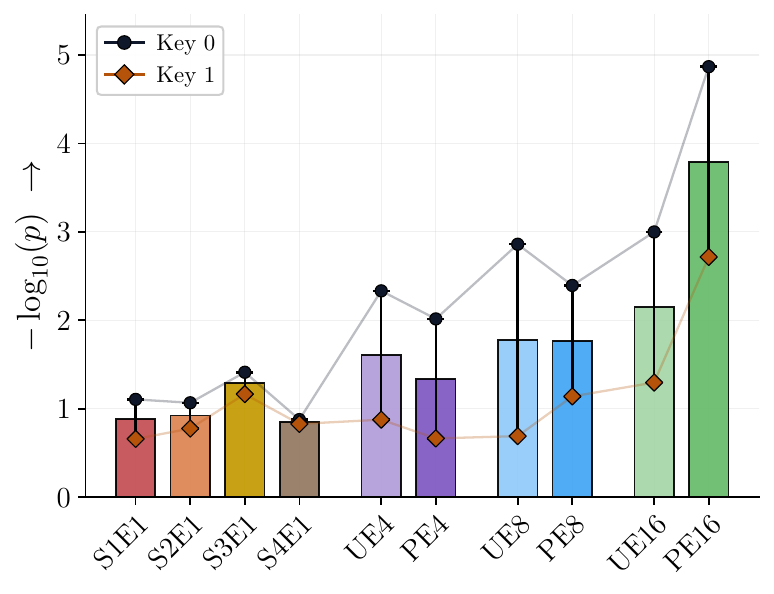}
    \caption{Pretraining CPT matched clean-twin false-probe null (left) and keyed signal (right) across the ten-schedule sweep at $F=2$, on the aligned unpacked detection surface, scored as $-\log_{10} p$ under the empirical-Gaussian reference.}
    \label{fig:appendix-phase2-cpt-aligned-gaussian}
\end{figure}

\begin{figure}[h!]
    \centering
    \includegraphics[width=0.49\textwidth]{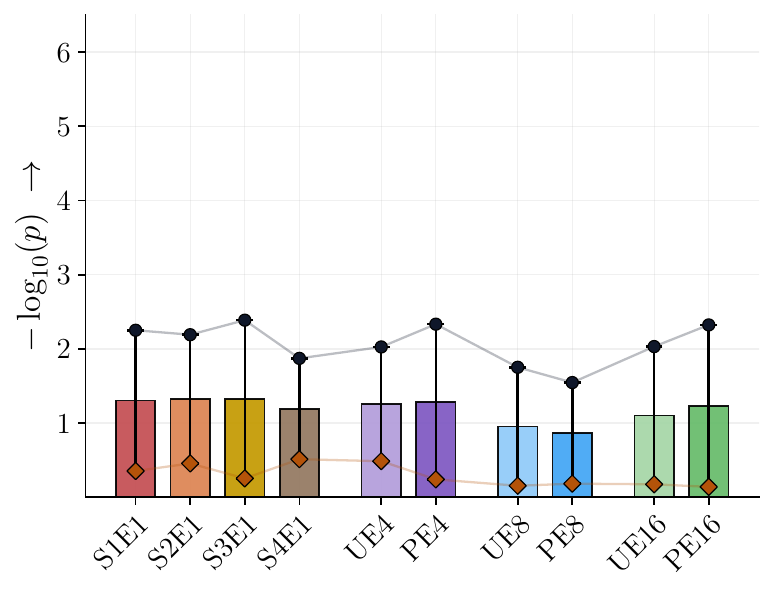}
    \hfill
    \includegraphics[width=0.49\textwidth]{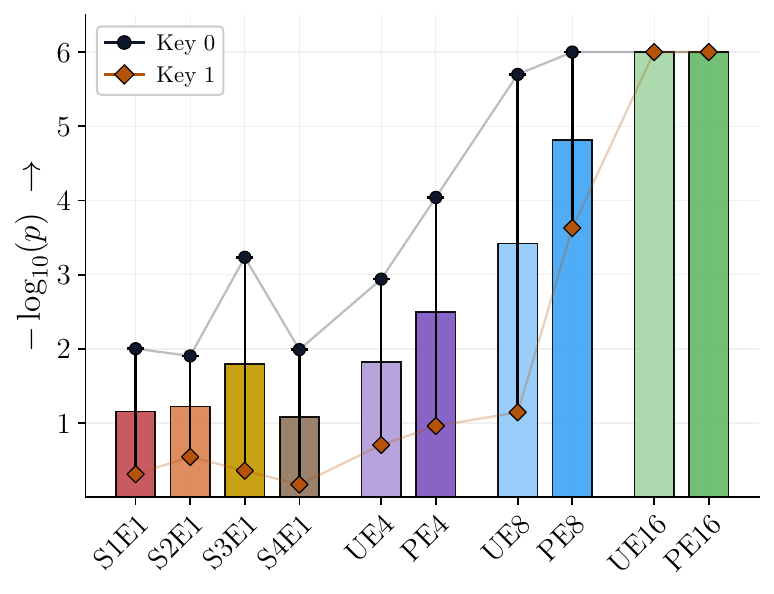}
    \caption{Pretraining CPT matched clean-twin false-probe null (left) and keyed signal (right) across the ten-schedule sweep at $F=2$, on the packed detection surface, scored as $-\log_{10} p$ under the empirical-exact null. The packed surface is a more permissive oracle baseline than the aligned surface.}
    \label{fig:appendix-phase2-cpt-packed-exact}
\end{figure}

\begin{figure}[h!]
    \centering
    \includegraphics[width=0.49\textwidth]{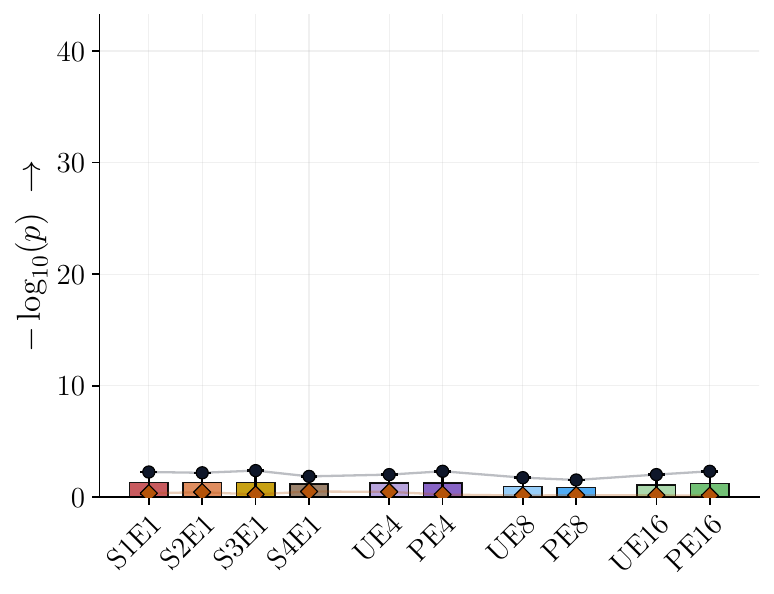}
    \hfill
    \includegraphics[width=0.49\textwidth]{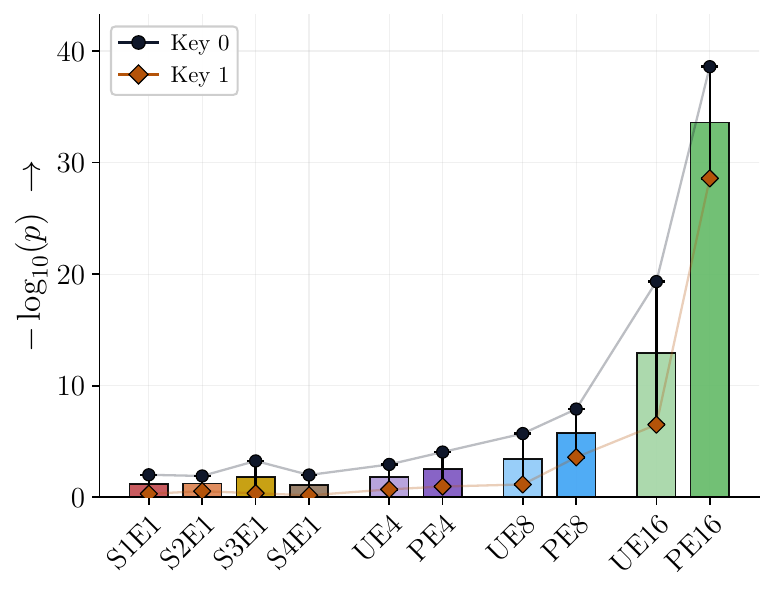}
    \caption{Pretraining CPT matched clean-twin false-probe null (left) and keyed signal (right) across the ten-schedule sweep at $F=2$, on the packed detection surface, scored as $-\log_{10} p$ under the empirical-Gaussian reference.}
    \label{fig:appendix-phase2-cpt-packed-gaussian}
\end{figure}

\begin{figure}[h!]
    \centering
    \includegraphics[width=0.49\textwidth]{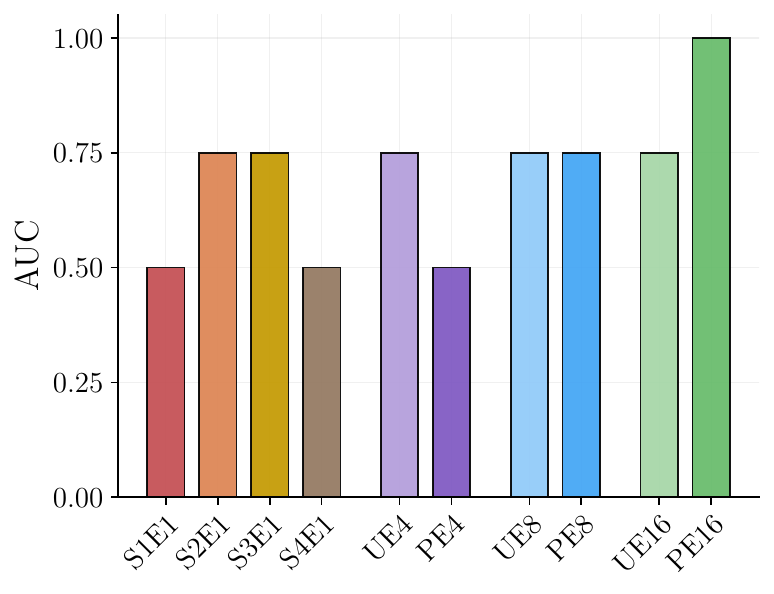}
    \hfill
    \includegraphics[width=0.49\textwidth]{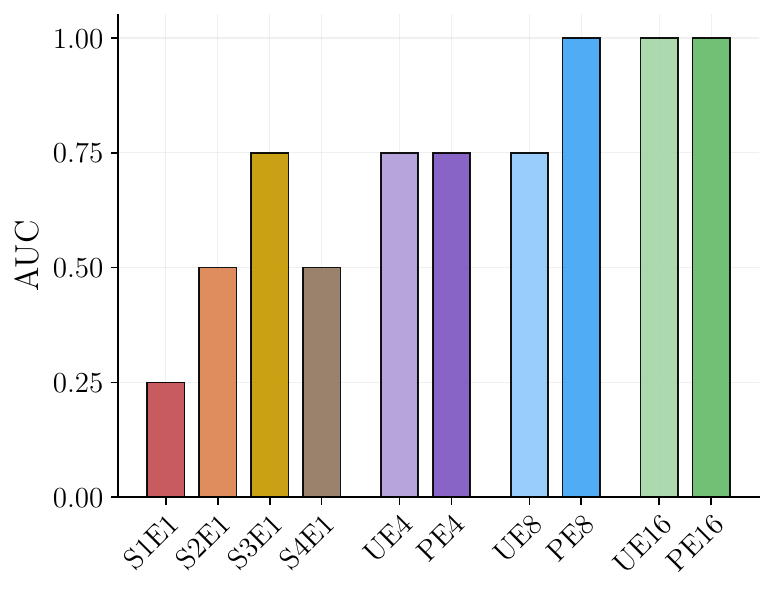}
    \caption{Pretraining CPT watermark whole-model DIA AUC across the ten-schedule sweep at $F=2$, on the aligned (left) and packed (right) detection surfaces, both scored against an empirical-exact null. Each schedule contributes $2{+}/2{-}$ whole-model trials, so AUC is coarse but tracks the keyed-signal ordering of \cref{fig:phase2-cpt-aligned-exact}.}
    \label{fig:appendix-phase2-cpt-dia-exact}
\end{figure}

\begin{figure}[h!]
    \centering
    \includegraphics[width=0.49\textwidth]{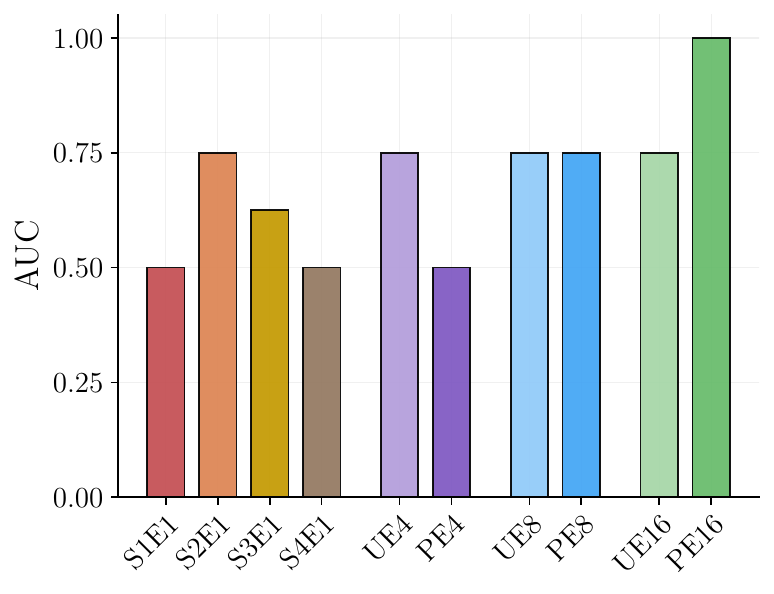}
    \hfill
    \includegraphics[width=0.49\textwidth]{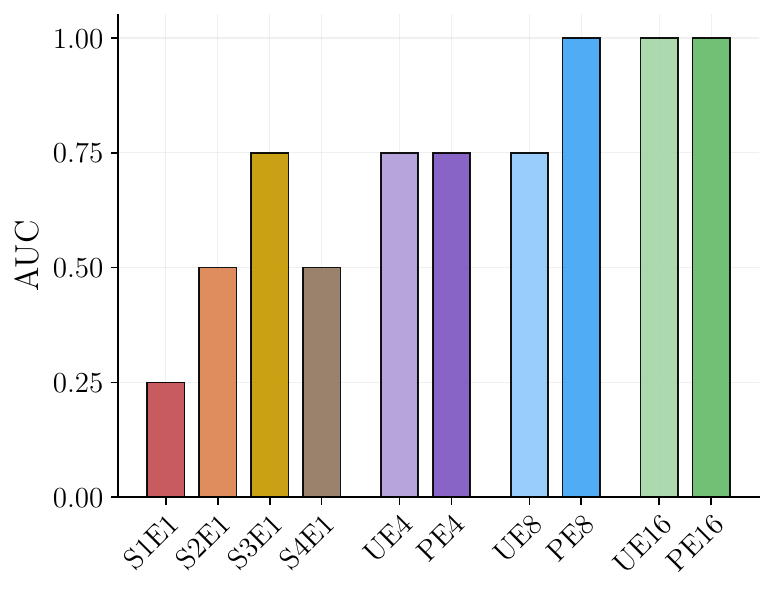}
    \caption{Pretraining CPT watermark whole-model DIA AUC across the ten-schedule sweep at $F=2$, on the aligned (left) and packed (right) detection surfaces, both scored against an empirical-Gaussian null. Each schedule contributes $2{+}/2{-}$ whole-model trials.}
    \label{fig:appendix-phase2-cpt-dia-gaussian}
\end{figure}

\Cref{fig:phase2-cpt-aligned-exact,fig:appendix-phase2-cpt-aligned-gaussian,fig:appendix-phase2-cpt-packed-exact,fig:appendix-phase2-cpt-packed-gaussian} carry the four CPT keyed/null pairs across the (aligned, packed) surface and (exact, Gaussian) $p$-value-type combinations. \Cref{fig:appendix-phase2-cpt-dia-exact,fig:appendix-phase2-cpt-dia-gaussian} carry the matching CPT watermark whole-model DIA AUC pairs, with aligned and packed surfaces shown side-by-side under each $p$-value type.

\subsection{Pretraining From-Scratch Companion Figures}\label{sec:appendix-phase2-scratch-companions}

\begin{figure}[h!]
    \centering
    \includegraphics[width=0.49\textwidth]{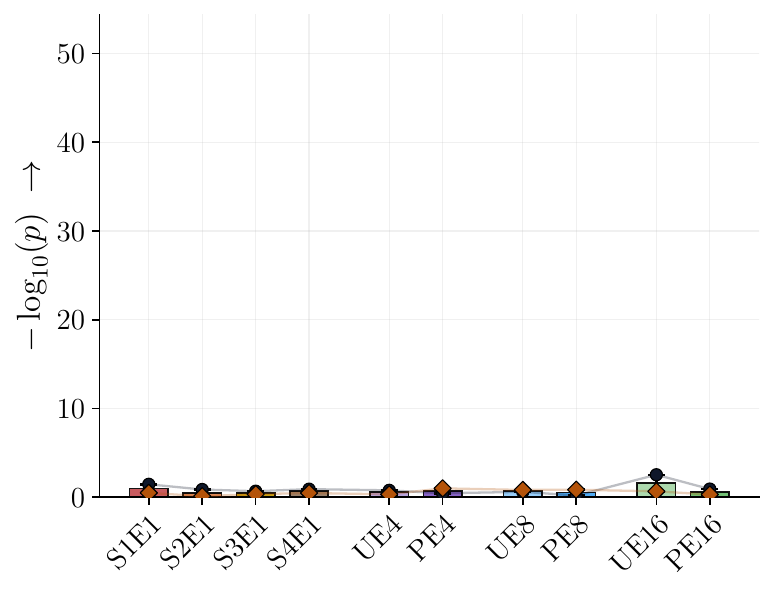}
    \hfill
    \includegraphics[width=0.49\textwidth]{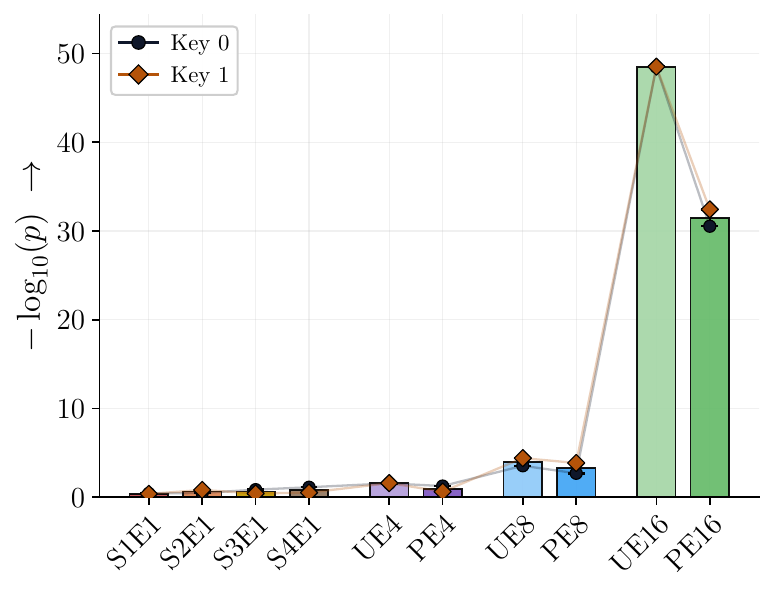}
    \caption{Pretraining from-scratch matched clean-twin false-probe null (left) and keyed signal (right) across the ten-schedule sweep at $F=2$, on the aligned unpacked detection surface, scored as $-\log_{10} p$ under the empirical-Gaussian reference.}
    \label{fig:appendix-phase2-scratch-aligned-gaussian}
\end{figure}

\begin{figure}[h!]
    \centering
    \includegraphics[width=0.49\textwidth]{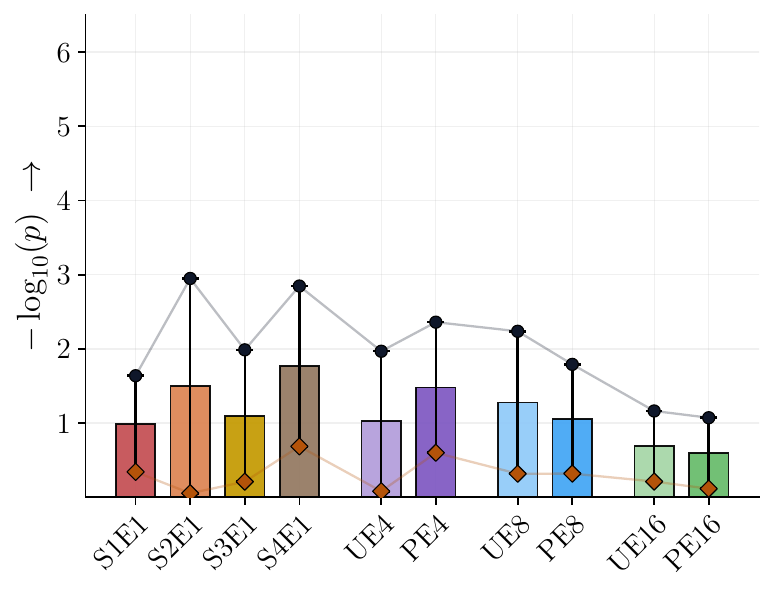}
    \hfill
    \includegraphics[width=0.49\textwidth]{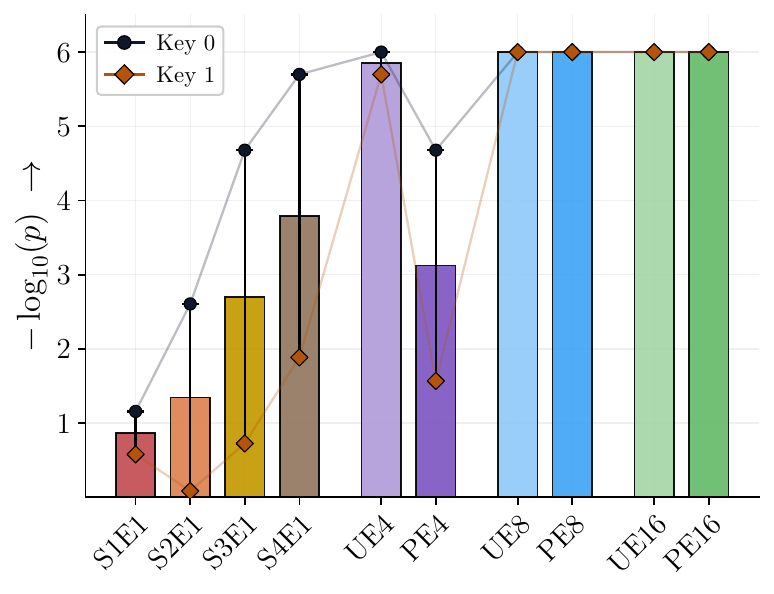}
    \caption{Pretraining from-scratch matched clean-twin false-probe null (left) and keyed signal (right) across the ten-schedule sweep at $F=2$, on the packed detection surface, scored as $-\log_{10} p$ under the empirical-exact null. The packed surface is a more permissive oracle baseline than the aligned surface.}
    \label{fig:appendix-phase2-scratch-packed-exact}
\end{figure}

\begin{figure}[h!]
    \centering
    \includegraphics[width=0.49\textwidth]{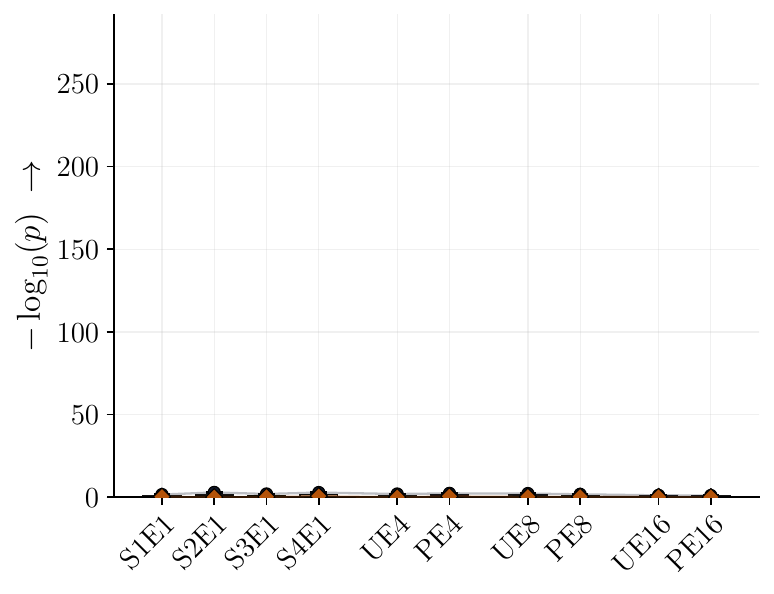}
    \hfill
    \includegraphics[width=0.49\textwidth]{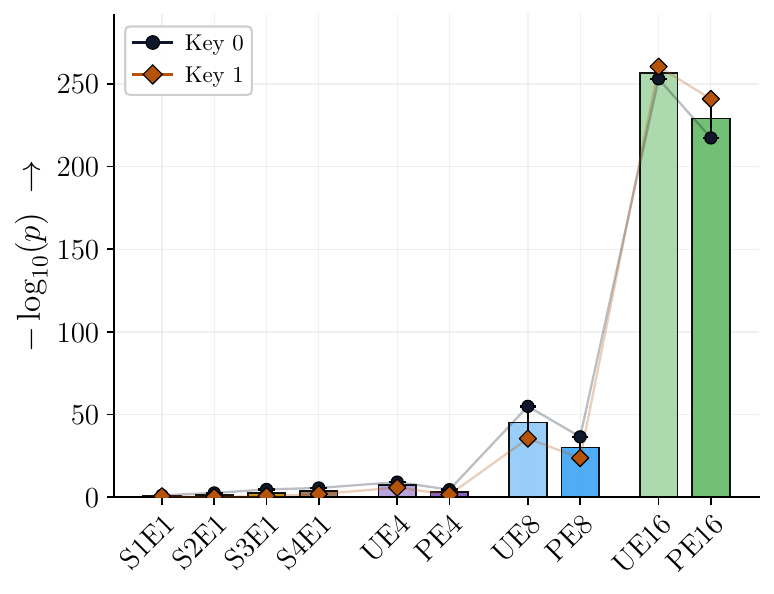}
    \caption{Pretraining from-scratch matched clean-twin false-probe null (left) and keyed signal (right) across the ten-schedule sweep at $F=2$, on the packed detection surface, scored as $-\log_{10} p$ under the empirical-Gaussian reference.}
    \label{fig:appendix-phase2-scratch-packed-gaussian}
\end{figure}

\begin{figure}[h!]
    \centering
    \includegraphics[width=0.49\textwidth]{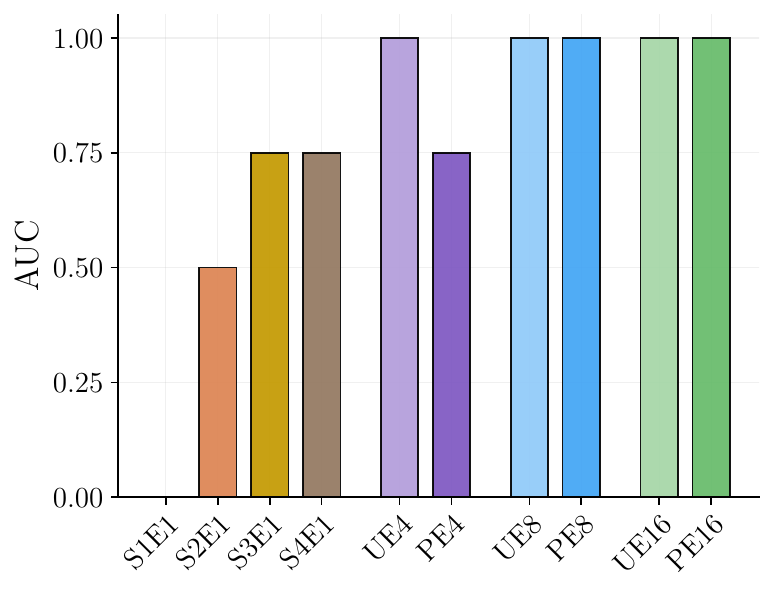}
    \hfill
    \includegraphics[width=0.49\textwidth]{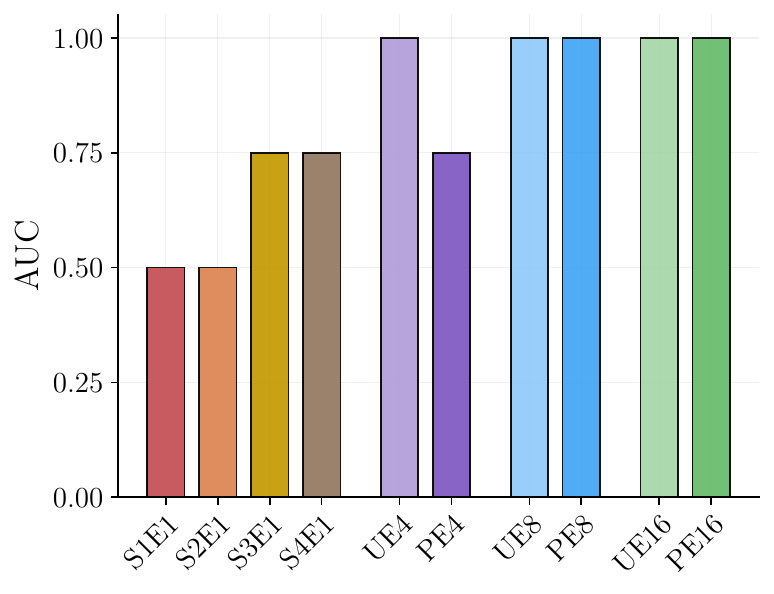}
    \caption{Pretraining from-scratch watermark whole-model DIA AUC across the ten-schedule sweep at $F=2$, on the aligned (left) and packed (right) detection surfaces, both scored against an empirical-exact null. Each schedule contributes $2{+}/2{-}$ whole-model trials. The from-scratch DIA recovers near-saturated AUC at the high-exposure schedules well before CPT does.}
    \label{fig:appendix-phase2-scratch-dia-exact}
\end{figure}

\begin{figure}[h!]
    \centering
    \includegraphics[width=0.49\textwidth]{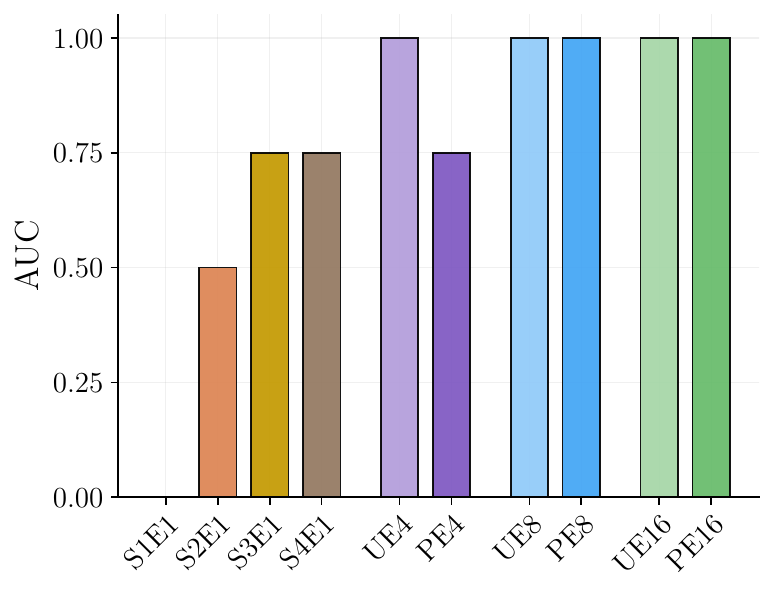}
    \hfill
    \includegraphics[width=0.49\textwidth]{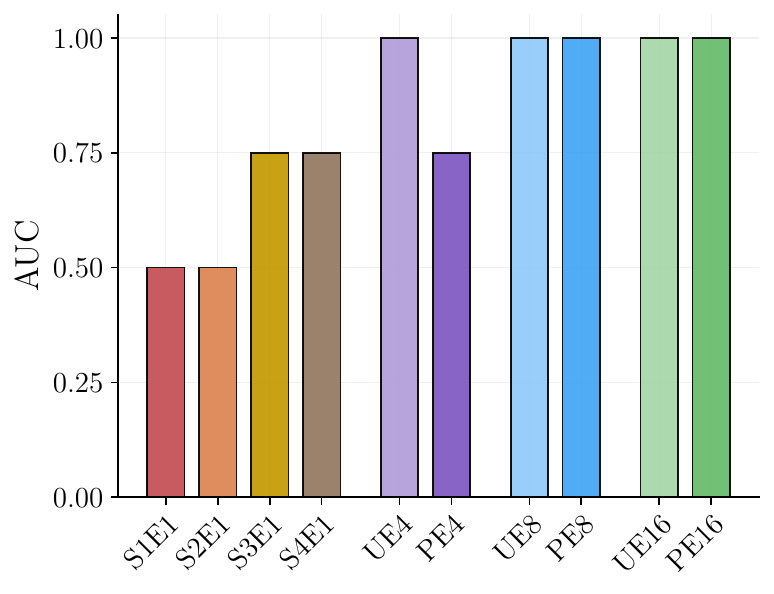}
    \caption{Pretraining from-scratch watermark whole-model DIA AUC across the ten-schedule sweep at $F=2$, on the aligned (left) and packed (right) detection surfaces, both scored against an empirical-Gaussian null. Each schedule contributes $2{+}/2{-}$ whole-model trials.}
    \label{fig:appendix-phase2-scratch-dia-gaussian}
\end{figure}

\Cref{fig:phase2-scratch-aligned-exact,fig:appendix-phase2-scratch-aligned-gaussian,fig:appendix-phase2-scratch-packed-exact,fig:appendix-phase2-scratch-packed-gaussian} carry the four from-scratch keyed/null pairs across the (aligned, packed) surface and (exact, Gaussian) $p$-value-type combinations. \Cref{fig:appendix-phase2-scratch-dia-exact,fig:appendix-phase2-scratch-dia-gaussian} carry the matching from-scratch watermark whole-model DIA AUC pairs, with aligned and packed surfaces shown side-by-side under each $p$-value type.

\subsection{Pretraining Schedule Summaries}\label{sec:appendix-phase2-provisional}

\Cref{tab:wm4memb-phase2-cpt-provisional,tab:wm4memb-phase2-scratch-provisional} provide compact per-schedule summaries combining the realized $\hat{E}$ with the aligned and packed watermark $-\log_{10} p$ values for both initialization regimes.

\input{auto_figures/wm4memb_phase2_cpt_provisional.tex}

\input{auto_figures/wm4memb_phase2_scratch_provisional.tex}

\subsection{Pretraining Realized Exposure and $\hat{E}$ Per-Schedule Readbacks}\label{sec:appendix-phase2-ehat}

\Cref{tab:wm4memb-phase2-cpt-exposure,tab:wm4memb-phase2-scratch-exposure} report the realized normalized exposure $\hat{E}/F$ per schedule for both pretraining initialization regimes, the direct realized counterpart to the idealized $E/F$ values in main-body \cref{tab:phase2-idealized-exposure}. \Cref{tab:wm4memb-phase2-cpt-ehat,tab:wm4memb-phase2-scratch-ehat} report the underlying realized $\hat{E}$ values. The corresponding idealized $E$ targets are the integers in the schedule names (column \texttt{$E$} of \cref{tab:phase2-idealized-exposure}).

\input{auto_figures/wm4memb_phase2_cpt_exposure_current.tex}

\input{auto_figures/wm4memb_phase2_scratch_exposure_current.tex}

\input{auto_figures/wm4memb_phase2_cpt_ehat_current.tex}

\input{auto_figures/wm4memb_phase2_scratch_ehat_current.tex}

\subsection{Pretraining Training Scale and Trial Geometry}\label{sec:appendix-phase2-context}

\Cref{tab:wm4memb-phase2-cpt-training-scale-context,tab:wm4memb-phase2-scratch-training-scale-context} report the per-schedule watermarked-token totals (mean, min--max across paired models) and the corresponding fraction of each model's 10.49B-token training budget for the CPT and from-scratch initialization regimes respectively. \Cref{tab:wm4memb-phase2-cpt-dia-context,tab:wm4memb-phase2-scratch-dia-context} report the per-schedule paired-model counts and the resulting $n_+ / n_-$ trial counts that drive each schedule's whole-model DIA AUC. Each schedule contributes $2 / 2$ positive/negative whole-model trials per init across both initialization regimes; \cref{sec:phase2} summarizes the extremal mass values.

\input{auto_figures/wm4memb_phase2_cpt_training_scale_context_current.tex}

\input{auto_figures/wm4memb_phase2_scratch_training_scale_context_current.tex}

\input{auto_figures/wm4memb_phase2_cpt_dia_context_current.tex}

\input{auto_figures/wm4memb_phase2_scratch_dia_context_current.tex}

\subsection{Pretraining Loss-Based and Reference-Model Row-Level MIA and Whole-Model DIA Baselines}\label{sec:appendix-phase2-baselines}

\Cref{tab:wm4memb-phase2-cpt-row-baselines,tab:wm4memb-phase2-cpt-dia-baselines,tab:wm4memb-phase2-scratch-row-baselines,tab:wm4memb-phase2-scratch-dia-baselines} report the source-fold loss-based and reference-model row-level MIA and whole-model DIA AUCs alongside the watermark detector's own AUCs (which are the same values that drive \cref{fig:appendix-phase2-cpt-dia-exact,fig:appendix-phase2-scratch-dia-exact}).

\input{auto_figures/wm4memb_phase2_cpt_row_baselines_current.tex}

\input{auto_figures/wm4memb_phase2_cpt_dia_baselines_current.tex}

\input{auto_figures/wm4memb_phase2_scratch_row_baselines_current.tex}

\input{auto_figures/wm4memb_phase2_scratch_dia_baselines_current.tex}

\end{document}

%% file: auto_figures/wm4memb_phase1_event_grid_fixed_dia_current.tex
\begin{table*}[h!]
\centering
\caption{Event-split finetuning: fold-level whole-model DIA AUC comparison. Entries marked N/A indicate statistics that are not estimable in the available cell geometry; for example, LiRA in F=2 cells lacks sufficient in-reference models.}
\label{tab:wm4memb-phase1-event-grid-fixed-dia-current}
\begin{tabular}{lcccccccc}
\hline
 & \multicolumn{2}{c|}{Watermark DIA} & \multicolumn{3}{c|}{Loss-based DIA} & \multicolumn{3}{c}{Ref-model DIA} \\
\cline{2-3}\cline{4-6}\cline{7-9}
Cell & Aligned & Packed & Raw-loss & Argmax & min-k$_{10}$ & rMIA-simple & rMIA & LiRA \\
\hline
(F2,E1) & 0.7500 & 1.0000 & 1.0000 & 1.0000 & 1.0000 & 1.0000 & 1.0000 & N/A \\
(F2,E2) & 1.0000 & 1.0000 & 1.0000 & 1.0000 & 1.0000 & 1.0000 & 1.0000 & N/A \\
(F2,E4) & 1.0000 & 1.0000 & 1.0000 & 1.0000 & 1.0000 & 1.0000 & 1.0000 & N/A \\
(F2,E8) & 1.0000 & 1.0000 & 1.0000 & 1.0000 & 1.0000 & 1.0000 & 1.0000 & N/A \\
(F2,E16) & 1.0000 & 1.0000 & 1.0000 & 1.0000 & 1.0000 & 1.0000 & 1.0000 & N/A \\
(F4,E1) & 0.5556 & 0.7292 & 1.0000 & 1.0000 & 1.0000 & 1.0000 & 1.0000 & 0.1181 \\
(F4,E2) & 0.7222 & 1.0000 & 1.0000 & 1.0000 & 1.0000 & 1.0000 & 1.0000 & 0.4861 \\
(F4,E4) & 0.7431 & 1.0000 & 1.0000 & 1.0000 & 1.0000 & 1.0000 & 1.0000 & 0.4514 \\
(F4,E8) & 1.0000 & 1.0000 & 1.0000 & 1.0000 & 1.0000 & 1.0000 & 1.0000 & 0.4514 \\
(F4,E16) & 1.0000 & 1.0000 & 1.0000 & 1.0000 & 1.0000 & 1.0000 & 1.0000 & 0.8056 \\
(F8,E1) & 0.5968 & 0.8815 & 1.0000 & 0.9987 & 1.0000 & 1.0000 & 1.0000 & 1.0000 \\
(F8,E2) & 0.7118 & 0.9974 & 1.0000 & 1.0000 & 1.0000 & 1.0000 & 1.0000 & 1.0000 \\
(F8,E4) & 0.9162 & 1.0000 & 1.0000 & 1.0000 & 1.0000 & 1.0000 & 1.0000 & 1.0000 \\
(F8,E8) & 1.0000 & 1.0000 & 1.0000 & 1.0000 & 1.0000 & 1.0000 & 1.0000 & 1.0000 \\
(F8,E16) & 1.0000 & 1.0000 & 1.0000 & 1.0000 & 1.0000 & 1.0000 & 1.0000 & 1.0000 \\
(F16,E1) & 0.5675 & 0.7062 & 0.9787 & 0.9702 & 0.9887 & 1.0000 & 1.0000 & 1.0000 \\
(F16,E2) & 0.6477 & 0.8949 & 1.0000 & 1.0000 & 1.0000 & 1.0000 & 1.0000 & 1.0000 \\
(F16,E4) & 0.8439 & 0.9999 & 1.0000 & 1.0000 & 1.0000 & 1.0000 & 1.0000 & 1.0000 \\
(F16,E8) & 0.9999 & 1.0000 & 1.0000 & 1.0000 & 1.0000 & 1.0000 & 1.0000 & 1.0000 \\
(F16,E16) & 1.0000 & 1.0000 & 1.0000 & 1.0000 & 1.0000 & 1.0000 & 1.0000 & 1.0000 \\
\end{tabular}
\end{table*}

%% file: auto_figures/wm4memb_phase1_event_grid_fixed_exposure_current.tex
\begin{table*}[h!]
\centering
\caption{Event-split finetuning: realized normalized exposure summary ($\hat{E}/F$).}
\label{tab:wm4memb-phase1-event-grid-fixed-exposure-current}
\begin{tabular}{lccccc}
\hline
  & $E=1$ & $E=2$ & $E=4$ & $E=8$ & $E=16$ \\
\hline
$F=2$ & 0.5612 & 1.1428 & 2.2880 & 4.3259 & 8.2775 \\
$F=4$ & 0.2944 & 0.5955 & 1.1603 & 2.2555 & 4.3436 \\
$F=8$ & 0.1490 & 0.2996 & 0.5883 & 1.1481 & 2.2133 \\
$F=16$ & 0.0837 & 0.1699 & 0.3310 & 0.6457 & 1.2488 \\
\end{tabular}
\end{table*}

%% file: auto_figures/wm4memb_phase1_event_grid_fixed_ehat_current.tex
\begin{table*}[h!]
\centering
\caption{Event-split finetuning: realized exposure summary ($\hat{E}$).}
\label{tab:wm4memb-phase1-event-grid-fixed-ehat-current}
\begin{tabular}{lccccc}
\hline
  & $E=1$ & $E=2$ & $E=4$ & $E=8$ & $E=16$ \\
\hline
$F=2$ & 1.1223 & 2.2857 & 4.5760 & 8.6519 & 16.5550 \\
$F=4$ & 1.1776 & 2.3818 & 4.6412 & 9.0218 & 17.3744 \\
$F=8$ & 1.1922 & 2.3970 & 4.7062 & 9.1846 & 17.7066 \\
$F=16$ & 1.3396 & 2.7188 & 5.2960 & 10.3307 & 19.9814 \\
\end{tabular}
\end{table*}

%% file: auto_figures/wm4memb_phase1_event_grid_fixed_training_scale_context_current.tex
\begin{table*}[h!]
\centering
\caption{Event-split finetuning: training scale context. Per-cell watermark token totals are relative to 131.1M train tokens per run. The percent columns report watermark-token share of total train tokens.}
\label{tab:wm4memb-phase1-event-fixed-training-scale-context}
\begin{tabular}{lccccc}
\hline
Cell & Target $(E/F)$ & WM tokens mean & WM tokens min-max & Mean \% & Range \% \\
\hline
(F2,E1) & 0.5000 & 0.5633M & 0.5367M-0.59M & 0.430\% & 0.409\%-0.450\% \\
(F2,E2) & 1.0000 & 1.143M & 1.082M-1.205M & 0.872\% & 0.825\%-0.919\% \\
(F2,E4) & 2.0000 & 2.288M & 2.171M-2.405M & 1.746\% & 1.657\%-1.835\% \\
(F2,E8) & 4.0000 & 4.318M & 4.294M-4.343M & 3.295\% & 3.276\%-3.313\% \\
(F2,E16) & 8.0000 & 8.294M & 8.239M-8.35M & 6.328\% & 6.286\%-6.370\% \\
(F4,E1) & 0.2500 & 0.2875M & 0.2212M-0.3687M & 0.219\% & 0.169\%-0.281\% \\
(F4,E2) & 0.5000 & 0.5797M & 0.463M-0.6883M & 0.442\% & 0.353\%-0.525\% \\
(F4,E4) & 1.0000 & 1.122M & 1.028M-1.307M & 0.856\% & 0.785\%-0.997\% \\
(F4,E8) & 2.0000 & 2.181M & 1.979M-2.339M & 1.664\% & 1.510\%-1.785\% \\
(F4,E16) & 4.0000 & 4.2M & 3.802M-4.49M & 3.204\% & 2.901\%-3.426\% \\
(F8,E1) & 0.1250 & 0.1417M & 0.09013M-0.2008M & 0.108\% & 0.069\%-0.153\% \\
(F8,E2) & 0.2500 & 0.286M & 0.2171M-0.3728M & 0.218\% & 0.166\%-0.284\% \\
(F8,E4) & 0.5000 & 0.5587M & 0.463M-0.7088M & 0.426\% & 0.353\%-0.541\% \\
(F8,E8) & 1.0000 & 1.09M & 0.9177M-1.397M & 0.832\% & 0.700\%-1.066\% \\
(F8,E16) & 2.0000 & 2.103M & 1.758M-2.454M & 1.604\% & 1.341\%-1.872\% \\
(F16,E1) & 0.0625 & 0.07084M & 0.02868M-0.1311M & 0.054\% & 0.022\%-0.100\% \\
(F16,E2) & 0.1250 & 0.143M & 0.09013M-0.2089M & 0.109\% & 0.069\%-0.159\% \\
(F16,E4) & 0.2500 & 0.2794M & 0.2212M-0.3605M & 0.213\% & 0.169\%-0.275\% \\
(F16,E8) & 0.5000 & 0.5452M & 0.4466M-0.7293M & 0.416\% & 0.341\%-0.556\% \\
(F16,E16) & 1.0000 & 1.051M & 0.8153M-1.377M & 0.802\% & 0.622\%-1.050\% \\
\end{tabular}
\end{table*}

%% file: auto_figures/wm4memb_phase1_event_grid_fixed_dia_context_current.tex
\begin{table*}[h!]
\centering
\caption{Event-split finetuning: model and watermark DIA trial geometry. The WM and clean model columns count target models per cell, and the WM DIA trial column counts the positive/negative pooled trials used for the watermark DIA AUC.}
\label{tab:wm4memb-phase1-event-fixed-dia-context}
\begin{tabular}{lccc}
\hline
Cell & WM models $n_+$ & Clean models $n_-$ & WM DIA trials $n_+ / n_-$ \\
\hline
(F2,E1) & 2 & 2 & 2 / 2 \\
(F2,E2) & 2 & 2 & 2 / 2 \\
(F2,E4) & 2 & 2 & 2 / 2 \\
(F2,E8) & 2 & 2 & 2 / 2 \\
(F2,E16) & 2 & 2 & 2 / 2 \\
(F4,E1) & 6 & 6 & 12 / 12 \\
(F4,E2) & 6 & 6 & 12 / 12 \\
(F4,E4) & 6 & 6 & 12 / 12 \\
(F4,E8) & 6 & 6 & 12 / 12 \\
(F4,E16) & 6 & 6 & 12 / 12 \\
(F8,E1) & 12 & 12 & 48 / 48 \\
(F8,E2) & 12 & 12 & 48 / 48 \\
(F8,E4) & 12 & 12 & 48 / 48 \\
(F8,E8) & 12 & 12 & 48 / 48 \\
(F8,E16) & 12 & 12 & 48 / 48 \\
(F16,E1) & 12 & 12 & 96 / 96 \\
(F16,E2) & 12 & 12 & 96 / 96 \\
(F16,E4) & 12 & 12 & 96 / 96 \\
(F16,E8) & 12 & 12 & 96 / 96 \\
(F16,E16) & 12 & 12 & 96 / 96 \\
\end{tabular}
\end{table*}

%% file: auto_figures/wm4memb_phase1_event_grid_fixed_row_baselines_current.tex
\begin{table*}[h!]
\centering
\caption{Event-split finetuning: row-level MIA AUC comparison. Entries marked N/A indicate statistics that are not estimable in the available cell geometry; for example, LiRA in F=2 cells lacks sufficient in-reference models.}
\label{tab:wm4memb-phase1-event-grid-fixed-row-baselines-current}
\begin{tabular}{lccccccc}
\hline
 & \multicolumn{1}{c|}{Watermark Readout} & \multicolumn{3}{c|}{Loss-based Row MIA} & \multicolumn{3}{c}{Ref-model Row MIA} \\
\cline{2-2}\cline{3-5}\cline{6-8}
Cell & WM $-\log_{10}(p)$ & Raw-loss & Argmax & min-k$_{10}$ & rMIA-simple & rMIA & LiRA \\
\hline
(F2,E1) & 0.1894 & 0.6096 & 0.6083 & 0.6292 & 1.0000 & 0.9993 & N/A \\
(F2,E2) & 0.4135 & 0.6988 & 0.6987 & 0.7318 & 1.0000 & 0.9993 & N/A \\
(F2,E4) & 1.4385 & 0.8521 & 0.8508 & 0.8991 & 1.0000 & 0.9993 & N/A \\
(F2,E8) & 11.8808 & 0.9744 & 0.9744 & 0.9707 & 0.9999 & 0.9993 & N/A \\
(F2,E16) & 86.6829 & 0.9832 & 0.9870 & 0.9683 & 0.9998 & 0.9993 & N/A \\
(F4,E1) & 0.3363 & 0.6041 & 0.6013 & 0.6237 & 0.9760 & 0.9805 & 0.8050 \\
(F4,E2) & 0.7380 & 0.6901 & 0.6896 & 0.7229 & 0.9985 & 0.9978 & 0.9165 \\
(F4,E4) & 1.3158 & 0.8360 & 0.8336 & 0.8813 & 1.0000 & 0.9993 & 0.9415 \\
(F4,E8) & 10.0393 & 0.9711 & 0.9691 & 0.9718 & 1.0000 & 0.9993 & 0.9416 \\
(F4,E16) & 63.7768 & 0.9745 & 0.9788 & 0.9608 & 0.9997 & 0.9992 & 0.9707 \\
(F8,E1) & 0.4487 & 0.6022 & 0.6010 & 0.6207 & 0.9645 & 0.9686 & 0.9668 \\
(F8,E2) & 0.5877 & 0.6852 & 0.6849 & 0.7163 & 0.9952 & 0.9946 & 0.9981 \\
(F8,E4) & 1.1352 & 0.8306 & 0.8284 & 0.8731 & 0.9992 & 0.9986 & 0.9996 \\
(F8,E8) & 6.8089 & 0.9564 & 0.9542 & 0.9552 & 0.9996 & 0.9990 & 0.9992 \\
(F8,E16) & 40.3926 & 0.9672 & 0.9717 & 0.9530 & 0.9995 & 0.9990 & 0.9991 \\
(F16,E1) & 0.5373 & 0.5975 & 0.5965 & 0.6157 & 0.9302 & 0.9341 & 0.9486 \\
(F16,E2) & 0.6138 & 0.6774 & 0.6758 & 0.7068 & 0.9771 & 0.9768 & 0.9911 \\
(F16,E4) & 1.1855 & 0.8148 & 0.8098 & 0.8471 & 0.9904 & 0.9897 & 0.9926 \\
(F16,E8) & 5.6955 & 0.9148 & 0.9166 & 0.9064 & 0.9903 & 0.9917 & 0.9898 \\
(F16,E16) & 20.9064 & 0.9294 & 0.9355 & 0.9132 & 0.9959 & 0.9962 & 0.9940 \\
\end{tabular}
\end{table*}

%% file: auto_figures/wm4memb_phase1_sks_exposure_current.tex
\begin{table*}[h!]
\centering
\caption{SKS finetuning: realized normalized exposure summary ($\hat{E}/F$).}
\label{tab:wm4memb-phase1-sks-exposure-current}
\begin{tabular}{lccccc}
\hline
  & $E=1$ & $E=2$ & $E=4$ & $E=8$ & $E=16$ \\
\hline
$F=2$ & 0.5557 & 1.1246 & 2.2800 & 4.3204 & 8.3766 \\
$F=4$ & 0.2755 & 0.5584 & 1.1609 & 2.2062 & 4.3422 \\
$F=8$ & 0.1549 & 0.2884 & 0.5772 & 1.1895 & 2.1984 \\
$F=16$ & 0.0911 & 0.1797 & 0.3477 & 0.6667 & 1.2930 \\
\end{tabular}
\end{table*}

%% file: auto_figures/wm4memb_phase1_sks_ehat_current.tex
\begin{table*}[h!]
\centering
\caption{SKS finetuning: realized exposure summary ($\hat{E}$).}
\label{tab:wm4memb-phase1-sks-ehat-current}
\begin{tabular}{lccccc}
\hline
  & $E=1$ & $E=2$ & $E=4$ & $E=8$ & $E=16$ \\
\hline
$F=2$ & 1.1114 & 2.2492 & 4.5600 & 8.6408 & 16.7533 \\
$F=4$ & 1.1021 & 2.2335 & 4.6437 & 8.8248 & 17.3686 \\
$F=8$ & 1.2388 & 2.3070 & 4.6175 & 9.5159 & 17.5870 \\
$F=16$ & 1.4583 & 2.8750 & 5.5625 & 10.6667 & 20.6875 \\
\end{tabular}
\end{table*}

%% file: auto_figures/wm4memb_phase1_sks_training_scale_context_current.tex
\begin{table*}[h!]
\centering
\caption{SKS finetuning: training scale context. Per-cell watermark token totals are relative to 131.1M train tokens per run. The percent columns report watermark-token share of total train tokens.}
\label{tab:wm4memb-phase1-sks-training-scale-context}
\begin{tabular}{lccccc}
\hline
Cell & Target $(E/F)$ & WM tokens mean & WM tokens min-max & Mean \% & Range \% \\
\hline
(F2,E1) & 0.5000 & 0.5562M & 0.4671M-0.6719M & 0.424\% & 0.356\%-0.513\% \\
(F2,E2) & 1.0000 & 1.133M & 0.9628M-1.168M & 0.865\% & 0.735\%-0.891\% \\
(F2,E4) & 2.0000 & 2.282M & 2.13M-2.417M & 1.741\% & 1.625\%-1.844\% \\
(F2,E8) & 4.0000 & 4.354M & 4.236M-4.449M & 3.322\% & 3.232\%-3.395\% \\
(F2,E16) & 8.0000 & 8.425M & 8.12M-8.677M & 6.428\% & 6.195\%-6.620\% \\
(F4,E1) & 0.2500 & 0.2709M & 0.2417M-0.3073M & 0.207\% & 0.184\%-0.234\% \\
(F4,E2) & 0.5000 & 0.548M & 0.5039M-0.6105M & 0.418\% & 0.384\%-0.466\% \\
(F4,E4) & 1.0000 & 1.142M & 1.028M-1.25M & 0.871\% & 0.785\%-0.953\% \\
(F4,E8) & 2.0000 & 2.165M & 2.081M-2.257M & 1.652\% & 1.588\%-1.722\% \\
(F4,E16) & 4.0000 & 4.26M & 3.974M-4.404M & 3.250\% & 3.032\%-3.360\% \\
(F8,E1) & 0.1250 & 0.1541M & 0.1393M-0.1762M & 0.118\% & 0.106\%-0.134\% \\
(F8,E2) & 0.2500 & 0.2868M & 0.2294M-0.3278M & 0.219\% & 0.175\%-0.250\% \\
(F8,E4) & 0.5000 & 0.5736M & 0.5408M-0.6023M & 0.438\% & 0.413\%-0.459\% \\
(F8,E8) & 1.0000 & 1.183M & 1.053M-1.25M & 0.903\% & 0.803\%-0.953\% \\
(F8,E16) & 2.0000 & 2.204M & 2.003M-2.397M & 1.681\% & 1.528\%-1.829\% \\
(F16,E1) & 0.0625 & 0.0717M & 0.05736M-0.09013M & 0.055\% & 0.044\%-0.069\% \\
(F16,E2) & 0.1250 & 0.1413M & 0.1024M-0.1762M & 0.108\% & 0.078\%-0.134\% \\
(F16,E4) & 0.2500 & 0.2735M & 0.2294M-0.3196M & 0.209\% & 0.175\%-0.244\% \\
(F16,E8) & 0.5000 & 0.5244M & 0.4712M-0.5777M & 0.400\% & 0.359\%-0.441\% \\
(F16,E16) & 1.0000 & 1.017M & 0.9341M-1.098M & 0.776\% & 0.713\%-0.838\% \\
\end{tabular}
\end{table*}

%% file: auto_figures/wm4memb_phase1_sks_dia_context_current.tex
\begin{table*}[h!]
\centering
\caption{SKS finetuning: model and watermark DIA trial geometry. The WM and clean model columns count target models per cell, and the WM DIA trial column counts the positive/negative pooled trials used for the watermark DIA AUC.}
\label{tab:wm4memb-phase1-sks-dia-context}
\begin{tabular}{lccc}
\hline
Cell & WM models $n_+$ & Clean models $n_-$ & WM DIA trials $n_+ / n_-$ \\
\hline
(F2,E1) & 4 & 4 & 4 / 4 \\
(F2,E2) & 4 & 4 & 4 / 4 \\
(F2,E4) & 4 & 4 & 4 / 4 \\
(F2,E8) & 4 & 4 & 4 / 4 \\
(F2,E16) & 4 & 4 & 4 / 4 \\
(F4,E1) & 4 & 4 & 4 / 4 \\
(F4,E2) & 4 & 4 & 4 / 4 \\
(F4,E4) & 4 & 4 & 4 / 4 \\
(F4,E8) & 4 & 4 & 4 / 4 \\
(F4,E16) & 4 & 4 & 4 / 4 \\
(F8,E1) & 4 & 4 & 4 / 4 \\
(F8,E2) & 4 & 4 & 4 / 4 \\
(F8,E4) & 4 & 4 & 4 / 4 \\
(F8,E8) & 4 & 4 & 4 / 4 \\
(F8,E16) & 4 & 4 & 4 / 4 \\
(F16,E1) & 4 & 4 & 4 / 4 \\
(F16,E2) & 4 & 4 & 4 / 4 \\
(F16,E4) & 4 & 4 & 4 / 4 \\
(F16,E8) & 4 & 4 & 4 / 4 \\
(F16,E16) & 4 & 4 & 4 / 4 \\
\end{tabular}
\end{table*}

%% file: auto_figures/wm4memb_phase1_sks_watermark_dia_current.tex
\begin{table*}[h!]
\centering
\caption{SKS finetuning: watermark whole-model DIA AUC summary.}
\label{tab:wm4memb-phase1-sks-watermark-dia-current}
\begin{tabular}{lcc}
\hline
Cell & Aligned Exact & Packed Exact \\
\hline
(F2,E1) & 0.6875 & 1.0000 \\
(F2,E2) & 0.7500 & 1.0000 \\
(F2,E4) & 0.9375 & 1.0000 \\
(F2,E8) & 1.0000 & 1.0000 \\
(F2,E16) & 1.0000 & 1.0000 \\
(F4,E1) & 0.5625 & 1.0000 \\
(F4,E2) & 0.5000 & 1.0000 \\
(F4,E4) & 0.9375 & 1.0000 \\
(F4,E8) & 1.0000 & 1.0000 \\
(F4,E16) & 1.0000 & 1.0000 \\
(F8,E1) & 0.6875 & 0.8125 \\
(F8,E2) & 0.6875 & 1.0000 \\
(F8,E4) & 0.8125 & 1.0000 \\
(F8,E8) & 1.0000 & 1.0000 \\
(F8,E16) & 1.0000 & 1.0000 \\
(F16,E1) & 0.6875 & 0.6875 \\
(F16,E2) & 0.7500 & 1.0000 \\
(F16,E4) & 0.8750 & 1.0000 \\
(F16,E8) & 1.0000 & 1.0000 \\
(F16,E16) & 1.0000 & 1.0000 \\
\end{tabular}
\end{table*}

%% file: auto_figures/wm4memb_phase2_cpt_provisional.tex
\begin{table*}[h!]
\centering
\caption{CPT pretraining: schedule summary.}
\label{tab:wm4memb-phase2-cpt-provisional}
\begin{tabular}{lccc}
\hline
Schedule & \shortstack{Realized\\($\hat{E}$)} & \shortstack{Aligned WM\\($-\log_{10}(p)$)} & \shortstack{Packed WM\\($-\log_{10}(p)$)} \\
\hline
(S1,E1) & 1.0000 & 0.8825 & 1.1630 \\
(S2,E1) & 1.0000 & 0.9225 & 1.2270 \\
(S3,E1) & 1.0000 & 1.2895 & 1.8080 \\
(S4,E1) & 1.0000 & 0.8535 & 1.0840 \\
(U,E4) & 3.8810 & 1.6050 & 1.8240 \\
(P,E4) & 4.0000 & 1.3405 & 2.5080 \\
(U,E8) & 7.8935 & 1.7755 & 3.4280 \\
(P,E8) & 8.0000 & 1.7670 & 5.7460 \\
(U,E16) & 16.2050 & 2.1480 & 12.9210 \\
(P,E16) & 16.0000 & 3.7915 & 33.5980 \\
\end{tabular}
\end{table*}

%% file: auto_figures/wm4memb_phase2_scratch_provisional.tex
\begin{table*}[h!]
\centering
\caption{From-scratch pretraining: schedule summary.}
\label{tab:wm4memb-phase2-scratch-provisional}
\begin{tabular}{lccc}
\hline
Schedule & \shortstack{Realized\\($\hat{E}$)} & \shortstack{Aligned WM\\($-\log_{10}(p)$)} & \shortstack{Packed WM\\($-\log_{10}(p)$)} \\
\hline
(S1,E1) & 1.0000 & 0.4025 & 0.8710 \\
(S2,E1) & 1.0000 & 0.6535 & 1.3460 \\
(S3,E1) & 1.0000 & 0.6415 & 2.7280 \\
(S4,E1) & 1.0000 & 0.8265 & 3.7550 \\
(U,E4) & 3.8895 & 1.5825 & 7.5010 \\
(P,E4) & 4.0000 & 0.9485 & 3.0750 \\
(U,E8) & 7.8115 & 3.9795 & 45.2420 \\
(P,E8) & 8.0000 & 3.2805 & 30.1350 \\
(U,E16) & 16.0740 & 48.4810 & 256.7420 \\
(P,E16) & 16.0000 & 31.4770 & 229.0900 \\
\end{tabular}
\end{table*}

%% file: auto_figures/wm4memb_phase2_cpt_exposure_current.tex
\begin{table*}[h!]
\centering
\caption{CPT pretraining: realized normalized exposure summary ($\hat{E}/F$).}
\label{tab:wm4memb-phase2-cpt-exposure}
\begin{tabular}{lc}
\hline
Schedule & \shortstack{Realized\\($\hat{E}/F$)} \\
\hline
(S1,E1) & 0.5000 \\
(S2,E1) & 0.5000 \\
(S3,E1) & 0.5000 \\
(S4,E1) & 0.5000 \\
(U,E4) & 1.9405 \\
(P,E4) & 2.0000 \\
(U,E8) & 3.9467 \\
(P,E8) & 4.0000 \\
(U,E16) & 8.1025 \\
(P,E16) & 8.0000 \\
\end{tabular}
\end{table*}

%% file: auto_figures/wm4memb_phase2_scratch_exposure_current.tex
\begin{table*}[h!]
\centering
\caption{From-scratch pretraining: realized normalized exposure summary ($\hat{E}/F$).}
\label{tab:wm4memb-phase2-scratch-exposure}
\begin{tabular}{lc}
\hline
Schedule & \shortstack{Realized\\($\hat{E}/F$)} \\
\hline
(S1,E1) & 0.5000 \\
(S2,E1) & 0.5000 \\
(S3,E1) & 0.5000 \\
(S4,E1) & 0.5000 \\
(U,E4) & 1.9447 \\
(P,E4) & 2.0000 \\
(U,E8) & 3.9057 \\
(P,E8) & 4.0000 \\
(U,E16) & 8.0370 \\
(P,E16) & 8.0000 \\
\end{tabular}
\end{table*}

%% file: auto_figures/wm4memb_phase2_cpt_ehat_current.tex
\begin{table*}[h!]
\centering
\caption{CPT pretraining: realized exposure summary ($\hat{E}$).}
\label{tab:wm4memb-phase2-cpt-ehat}
\begin{tabular}{lc}
\hline
Schedule & \shortstack{Realized\\($\hat{E}$)} \\
\hline
(S1,E1) & 1.0000 \\
(S2,E1) & 1.0000 \\
(S3,E1) & 1.0000 \\
(S4,E1) & 1.0000 \\
(U,E4) & 3.8810 \\
(P,E4) & 4.0000 \\
(U,E8) & 7.8935 \\
(P,E8) & 8.0000 \\
(U,E16) & 16.2050 \\
(P,E16) & 16.0000 \\
\end{tabular}
\end{table*}

%% file: auto_figures/wm4memb_phase2_scratch_ehat_current.tex
\begin{table*}[h!]
\centering
\caption{From-scratch pretraining: realized exposure summary ($\hat{E}$).}
\label{tab:wm4memb-phase2-scratch-ehat}
\begin{tabular}{lc}
\hline
Schedule & \shortstack{Realized\\($\hat{E}$)} \\
\hline
(S1,E1) & 1.0000 \\
(S2,E1) & 1.0000 \\
(S3,E1) & 1.0000 \\
(S4,E1) & 1.0000 \\
(U,E4) & 3.8895 \\
(P,E4) & 4.0000 \\
(U,E8) & 7.8115 \\
(P,E8) & 8.0000 \\
(U,E16) & 16.0740 \\
(P,E16) & 16.0000 \\
\end{tabular}
\end{table*}

%% file: auto_figures/wm4memb_phase2_cpt_training_scale_context_current.tex
\begin{table*}[h!]
\centering
\caption{CPT pretraining: training scale context. Per-schedule watermark token totals are relative to 10.49B train tokens per run. The percent columns report watermark-token share of total train tokens.}
\label{tab:wm4memb-phase2-cpt-training-scale-context}
\begin{tabular}{lcccc}
\hline
Schedule & WM tokens seen & Realized $\hat{E}$ & Mean \% & Range \% \\
\hline
(S1,E1) & 0.50M & 1.000 & 0.005\% & 0.005\%-0.005\% \\
(S2,E1) & 0.50M & 1.000 & 0.005\% & 0.005\%-0.005\% \\
(S3,E1) & 0.50M & 1.000 & 0.005\% & 0.005\%-0.005\% \\
(S4,E1) & 0.50M & 1.000 & 0.005\% & 0.005\%-0.005\% \\
(U,E4) & 1.94M-1.94M & 3.877-3.885 & 0.019\% & 0.019\%-0.019\% \\
(P,E4) & 2.00M & 4.000 & 0.019\% & 0.019\%-0.019\% \\
(U,E8) & 3.88M-4.01M & 7.762-8.025 & 0.038\% & 0.037\%-0.038\% \\
(P,E8) & 4.00M & 8.000 & 0.038\% & 0.038\%-0.038\% \\
(U,E16) & 8.10M-8.10M & 16.197-16.213 & 0.077\% & 0.077\%-0.077\% \\
(P,E16) & 8.00M & 16.000 & 0.076\% & 0.076\%-0.076\% \\
\end{tabular}
\end{table*}

%% file: auto_figures/wm4memb_phase2_scratch_training_scale_context_current.tex
\begin{table*}[h!]
\centering
\caption{From-scratch pretraining: training scale context. Per-schedule watermark token totals are relative to 10.49B train tokens per run. The percent columns report watermark-token share of total train tokens.}
\label{tab:wm4memb-phase2-scratch-training-scale-context}
\begin{tabular}{lcccc}
\hline
Schedule & WM tokens seen & Realized $\hat{E}$ & Mean \% & Range \% \\
\hline
(S1,E1) & 0.500M & 1.000 & 0.005\% & 0.005\%-0.005\% \\
(S2,E1) & 0.500M & 1.000 & 0.005\% & 0.005\%-0.005\% \\
(S3,E1) & 0.500M & 1.000 & 0.005\% & 0.005\%-0.005\% \\
(S4,E1) & 0.500M & 1.000 & 0.005\% & 0.005\%-0.005\% \\
(U,E4) & 1.938-1.950M & 3.877-3.902 & 0.019\% & 0.018\%-0.019\% \\
(P,E4) & 1.999M & 4.000 & 0.019\% & 0.019\%-0.019\% \\
(U,E8) & 3.827-3.982M & 7.656-7.967 & 0.037\% & 0.036\%-0.038\% \\
(P,E8) & 3.999M & 8.000 & 0.038\% & 0.038\%-0.038\% \\
(U,E16) & 7.973-8.096M & 15.951-16.197 & 0.077\% & 0.076\%-0.077\% \\
(P,E16) & 7.997M & 16.000 & 0.076\% & 0.076\%-0.076\% \\
\end{tabular}
\end{table*}

%% file: auto_figures/wm4memb_phase2_cpt_dia_context_current.tex
\begin{table*}[h!]
\centering
\caption{CPT pretraining: model and watermark DIA trial geometry. In these pretraining cells, model counts and watermark DIA trial counts coincide because each target contributes one positive or negative whole-model trial to the pooled AUC.}
\label{tab:wm4memb-phase2-cpt-dia-context}
\begin{tabular}{lccc}
\hline
Schedule & WM models $n_+$ & Clean models $n_-$ & WM DIA trials $n_+ / n_-$ \\
\hline
(S1,E1) & 2 & 2 & 2 / 2 \\
(S2,E1) & 2 & 2 & 2 / 2 \\
(S3,E1) & 2 & 2 & 2 / 2 \\
(S4,E1) & 2 & 2 & 2 / 2 \\
(U,E4) & 2 & 2 & 2 / 2 \\
(P,E4) & 2 & 2 & 2 / 2 \\
(U,E8) & 2 & 2 & 2 / 2 \\
(P,E8) & 2 & 2 & 2 / 2 \\
(U,E16) & 2 & 2 & 2 / 2 \\
(P,E16) & 2 & 2 & 2 / 2 \\
\end{tabular}
\end{table*}

%% file: auto_figures/wm4memb_phase2_scratch_dia_context_current.tex
\begin{table*}[h!]
\centering
\caption{From-scratch pretraining: model and watermark DIA trial geometry. In these pretraining cells, model counts and watermark DIA trial counts coincide because each target contributes one positive or negative whole-model trial to the pooled AUC.}
\label{tab:wm4memb-phase2-scratch-dia-context}
\begin{tabular}{lccc}
\hline
Schedule & WM models $n_+$ & Clean models $n_-$ & WM DIA trials $n_+ / n_-$ \\
\hline
(S1,E1) & 2 & 2 & 2 / 2 \\
(S2,E1) & 2 & 2 & 2 / 2 \\
(S3,E1) & 2 & 2 & 2 / 2 \\
(S4,E1) & 2 & 2 & 2 / 2 \\
(U,E4) & 2 & 2 & 2 / 2 \\
(P,E4) & 2 & 2 & 2 / 2 \\
(U,E8) & 2 & 2 & 2 / 2 \\
(P,E8) & 2 & 2 & 2 / 2 \\
(U,E16) & 2 & 2 & 2 / 2 \\
(P,E16) & 2 & 2 & 2 / 2 \\
\end{tabular}
\end{table*}

%% file: auto_figures/wm4memb_phase2_cpt_row_baselines_current.tex
\begin{table*}[h!]
\centering
\caption{CPT pretraining: row-level MIA AUC comparison.}
\label{tab:wm4memb-phase2-cpt-row-baselines}
\begin{tabular}{lccccc}
\hline
 & \multicolumn{1}{c|}{Watermark Readout} & \multicolumn{4}{c}{Loss-based Row MIA} \\
\cline{2-2}\cline{3-6}
Schedule & WM $-\log_{10}(p_{\mathrm{exact}})$ & Raw-loss & Argmax & min-k$_{10}$ & zlib \\
\hline
(S1,E1) & 0.8810 & 0.5029 & 0.5021 & 0.5036 & 0.5023 \\
(S2,E1) & 0.9195 & 0.5165 & 0.5142 & 0.5206 & 0.5127 \\
(S3,E1) & 1.2870 & 0.5291 & 0.5247 & 0.5367 & 0.5225 \\
(S4,E1) & 0.8515 & 0.5079 & 0.5059 & 0.5104 & 0.5062 \\
(U,E4) & 1.6045 & 0.5497 & 0.5457 & 0.5612 & 0.5387 \\
(P,E4) & 1.3390 & 0.5765 & 0.5700 & 0.5939 & 0.5601 \\
(U,E8) & 1.7695 & 0.6104 & 0.6015 & 0.6333 & 0.5893 \\
(P,E8) & 1.7695 & 0.6610 & 0.6521 & 0.6932 & 0.6328 \\
(U,E16) & 2.1415 & 0.7197 & 0.7112 & 0.7540 & 0.6906 \\
(P,E16) & 3.7765 & 0.8163 & 0.8117 & 0.8636 & 0.7884 \\
\end{tabular}
\end{table*}

%% file: auto_figures/wm4memb_phase2_cpt_dia_baselines_current.tex
\begin{table*}[h!]
\centering
\caption{CPT pretraining: fold-level whole-model DIA AUC comparison. Each cell contributes $2+/2-$ whole-model trials.}
\label{tab:wm4memb-phase2-cpt-dia-baselines}
\begin{tabular}{lcccccc}
\hline
 & \multicolumn{2}{c|}{Watermark DIA} & \multicolumn{4}{c}{Loss-based DIA} \\
\cline{2-3}\cline{4-7}
Schedule & Aligned & Packed & Raw-loss & Argmax & min-k$_{10}$ & zlib \\
\hline
(S1,E1) & 0.5000 & 0.2500 & 0.7500 & 0.7500 & 0.7500 & 0.7500 \\
(S2,E1) & 0.7500 & 0.5000 & 1.0000 & 1.0000 & 1.0000 & 1.0000 \\
(S3,E1) & 0.7500 & 0.7500 & 1.0000 & 1.0000 & 1.0000 & 1.0000 \\
(S4,E1) & 0.5000 & 0.5000 & 1.0000 & 0.7500 & 1.0000 & 0.7500 \\
(U,E4) & 0.7500 & 0.7500 & 1.0000 & 1.0000 & 1.0000 & 1.0000 \\
(P,E4) & 0.5000 & 0.7500 & 1.0000 & 1.0000 & 1.0000 & 1.0000 \\
(U,E8) & 0.7500 & 0.7500 & 1.0000 & 1.0000 & 1.0000 & 1.0000 \\
(P,E8) & 0.7500 & 1.0000 & 1.0000 & 1.0000 & 1.0000 & 1.0000 \\
(U,E16) & 0.7500 & 1.0000 & 1.0000 & 1.0000 & 1.0000 & 1.0000 \\
(P,E16) & 1.0000 & 1.0000 & 1.0000 & 1.0000 & 1.0000 & 1.0000 \\
\end{tabular}
\end{table*}

%% file: auto_figures/wm4memb_phase2_scratch_row_baselines_current.tex
\begin{table*}[h!]
\centering
\caption{From-scratch pretraining: row-level MIA AUC comparison.}
\label{tab:wm4memb-phase2-scratch-row-baselines}
\begin{tabular}{lccccc}
\hline
 & \multicolumn{1}{c|}{Watermark Readout} & \multicolumn{4}{c}{Loss-based Row MIA} \\
\cline{2-2}\cline{3-6}
Schedule & WM $-\log_{10}(p_{\mathrm{exact}})$ & Raw-loss & Argmax & min-k$_{10}$ & zlib \\
\hline
(S1,E1) & 0.4015 & 0.5028 & 0.5034 & 0.5064 & 0.5019 \\
(S2,E1) & 0.6520 & 0.5204 & 0.5130 & 0.5245 & 0.5141 \\
(S3,E1) & 0.6395 & 0.5823 & 0.5648 & 0.6058 & 0.5628 \\
(S4,E1) & 0.8250 & 0.6025 & 0.5957 & 0.6517 & 0.5797 \\
(U,E4) & 1.5800 & 0.6846 & 0.6621 & 0.7379 & 0.6525 \\
(P,E4) & 0.9470 & 0.6429 & 0.6043 & 0.6794 & 0.6106 \\
(U,E8) & 4.0410 & 0.8418 & 0.8209 & 0.8848 & 0.8122 \\
(P,E8) & 3.2730 & 0.8472 & 0.8072 & 0.8937 & 0.8131 \\
(U,E16) & 6.0000 & 0.9744 & 0.9768 & 0.9602 & 0.9715 \\
(P,E16) & 6.0000 & 0.9792 & 0.9764 & 0.9711 & 0.9765 \\
\end{tabular}
\end{table*}

%% file: auto_figures/wm4memb_phase2_scratch_dia_baselines_current.tex
\begin{table*}[h!]
\centering
\caption{From-scratch pretraining: fold-level whole-model DIA AUC comparison. Each cell contributes $2+/2-$ whole-model trials.}
\label{tab:wm4memb-phase2-scratch-dia-baselines}
\begin{tabular}{lcccccc}
\hline
 & \multicolumn{2}{c|}{Watermark DIA} & \multicolumn{4}{c}{Loss-based DIA} \\
\cline{2-3}\cline{4-7}
Schedule & Aligned & Packed & Raw-loss & Argmax & min-k$_{10}$ & zlib \\
\hline
(S1,E1) & 0.0000 & 0.5000 & 0.7500 & 1.0000 & 1.0000 & 0.5000 \\
(S2,E1) & 0.5000 & 0.5000 & 1.0000 & 1.0000 & 1.0000 & 1.0000 \\
(S3,E1) & 0.7500 & 0.7500 & 1.0000 & 1.0000 & 1.0000 & 1.0000 \\
(S4,E1) & 0.7500 & 0.7500 & 1.0000 & 1.0000 & 1.0000 & 1.0000 \\
(U,E4) & 1.0000 & 1.0000 & 1.0000 & 1.0000 & 1.0000 & 1.0000 \\
(P,E4) & 0.7500 & 0.7500 & 1.0000 & 1.0000 & 1.0000 & 1.0000 \\
(U,E8) & 1.0000 & 1.0000 & 1.0000 & 1.0000 & 1.0000 & 1.0000 \\
(P,E8) & 1.0000 & 1.0000 & 1.0000 & 1.0000 & 1.0000 & 1.0000 \\
(U,E16) & 1.0000 & 1.0000 & 1.0000 & 1.0000 & 1.0000 & 1.0000 \\
(P,E16) & 1.0000 & 1.0000 & 1.0000 & 1.0000 & 1.0000 & 1.0000 \\
\end{tabular}
\end{table*}